\documentclass{article}

\usepackage{arxiv}

\usepackage[utf8]{inputenc} 
\usepackage[T1]{fontenc}    
\usepackage{hyperref}       
\usepackage{url}            
\usepackage{booktabs}       
\usepackage{amsfonts}       
\usepackage{nicefrac}       
\usepackage{microtype}      
\usepackage{lipsum}
\usepackage{graphicx}
\usepackage{amsmath}
\usepackage{graphicx} 
\usepackage[table]{xcolor}
\graphicspath{ {./} }

\title{HIA-GAT: A Heterogeneous Interaction-Aware Graph Attention Network for Frame-Level Traffic Conflict Risk Prediction on Freeways}

\author{
 Mahshid Malazizi \\
 Center for Urban Informatics and Progress\\
  University of Tennessee at Chattanooga\\
  Chattanooga, TN, USA, 37403\\
  \texttt{ntv774@mocs.utc.edu} \\
   \And
 Seyedmehdi Khaleghian \\
 Center for Urban Informatics and Progress\\
  University of Tennessee at Chattanooga\\
  Chattanooga, TN, USA, 37403\\
  \texttt{mehdi-khaleghian@utc.edu} \\
  \And
  Mina Sartipi \\
 Center for Urban Informatics and Progress\\
  University of Tennessee at Chattanooga\\
  Chattanooga, TN, USA, 37403\\
  \texttt{Mina-Sartipi@utc.edu} \\
  \And
   Toru Hirano \\
  DENSO International America, Inc.\\
  Southfield, MI, USA \\
  \And
    Yunfei Xu \\
    DENSO International America, Inc.\\
  Southfield, MI, USA \\
  \And
   Hoang H. Nguyen \\
  Center for Urban Informatics and Progress\\
  University of Tennessee at Chattanooga\\
  Chattanooga, TN, USA, 37403\\
  \texttt{huuhoang-nguyen@utc.edu} }

\begin{document}
\maketitle
\begin{abstract}
This paper formulates frame-level freeway risk assessment as a multi-agent scene (graph-level) binary classification problem, where each video/trajectory frame is labeled risky if any TTC- or PET-based conflict violates a specified severity threshold. We construct a relation-aware graph per frame with vehicles as nodes and two interaction types as edges—same-lane (longitudinal) and adjacent-lane (lateral)—augmented with physics-informed edge features aligned to rear-end and lane-change conflict mechanisms. Building on a structured benchmarking suite of non-graph models and graph baselines, we propose HIA-GAT, a dual-stream heterogeneous graph attention network that processes longitudinal and lateral interactions through dedicated attention pathways and fuses them via a conflict-type-aware gating mechanism with event-level gate supervision derived from SSM conflict attribution. Experiments on the NGSIM I-80 and US-101 freeway datasets across nine TTC/PET threshold configurations show that HIA-GAT achieves the best average risk-ranking performance (AUC 0.835 on I-80 and 0.867 on US-101), with the largest gains on PET-only (lane-change) settings where relational structure is essential. Beyond accuracy, the learned gate provides interpretable per-vehicle attribution of dominant conflict type, supporting actionable, real-time freeway safety monitoring. We show that graph structure is critical for modeling lateral conflict risk, while longitudinal risk can often be captured by non-relational aggregation.
 
\end{abstract}


\section{I Introduction}

Traffic safety analysis at intersections and highway facilities has traditionally relied on historical crash records. Although essential for long-term planning, crash-based methods are reactive and constrained by sparse, delayed observations \cite{fhwa_using_safety_analyses_2025}. To support proactive evaluation, surrogate safety measures (SSMs), particularly Time-to-Collision (TTC) and Post-Encroachment Time (PET), are widely used to quantify conflict severity and near-miss dynamics before crashes occur \cite{hayward1972near}. However, real-time safety assessment at complex facilities remains challenging.
Roadside video sensing and computer vision now provide a scalable, cost-effective way to extract contextual and kinematic information for multiple road users \cite{chitraranjan2025vision}. Yet converting noisy multi-agent trajectories into reliable frame-level safety indicators remains difficult because SSMs are defined on pairwise interactions, while real scenes contain many simultaneous interactions that must be summarized at the scene level. Existing conflict-based studies mainly focus on intersections and specific interaction types \cite{singh2024conflict,bonela2022review,chaudhari2021exploring}, while broader SSM research emphasizes thresholds, crash-outcome links, contributing factors, and validation against crash data \cite{singh2024conflict,bataineh2025evaluating,hasain2024proposing,dimitrijevic2022short}. This leaves a gap in video-compatible, real-time, frame-level risk assessment for scenes with multiple interacting agents.

This gap is especially consequential for highway environments, which have been comparatively less emphasized in conflict-based risk evaluation despite exhibiting distinct traffic dynamics relative to intersections. On freeways and expressways, the dominant conflict mechanisms are typically rear-end interactions and lane-change/sideswipe conflicts \cite{qi2020modified,zhao2025traffic}. Empirical findings further suggest that lane-changing behavior and its associated risk are strongly modulated by longitudinal spacing to the relevant preceding vehicle following the maneuver \cite{xiang2024research}. These characteristics motivate a modeling approach that (i) can represent many concurrent interactions, (ii) can distinguish conflict mechanisms (longitudinal vs.\ lateral), and (iii) can deliver frame-resolved risk predictions suitable for continuous monitoring.

To address these limitations, we develop a comprehensive \emph{frame-level} traffic risk assessment framework for highway scenes grounded in TTC and PET. We organize our evaluation into three tiers of increasing architectural complexity: \emph{(1) non-graph baselines} that operate on frame-level aggregates, \emph{(2) GNN baselines without edge features} to isolate the contribution of relational structure, and \emph{(3) a single-stream GNN baseline (HomoGAT)} that incorporates SSM-relevant information through unified message passing under multiple SSM threshold settings and their combinations. Building on these baselines, we propose \textbf{HIA-GAT}, a heterogeneous interaction-aware graph attention network that decomposes vehicle interactions into conflict-type-specific message-passing streams---\emph{longitudinal} and \emph{lateral}---and fuses them via a gated mechanism informed by physics-grounded edge features. This design enables principled, frame-level traffic risk classification that is aligned with the underlying mechanisms captured by TTC and PET, while remaining compatible with video-derived multi-agent trajectories.

\noindent In summary, this paper makes the following contributions:
\begin{itemize}
    \item We formalize \emph{frame-level} highway risk assessment as a multi-agent scene classification problem using TTC and PET to derive operational risk labels from trajectory data.
    \item We provide a structured benchmarking suite spanning non-graph models, topology-only GNNs, and a single-stream attention-based GNN baseline (HomoGAT) under multiple SSM threshold configurations and their combinations.
    \item We introduce HIA-GAT, a heterogeneous interaction-aware architecture with conflict-type-specific message passing and gated fusion using physics-grounded edge features, enabling mechanism-aligned and interpretable frame-level risk prediction.
\end{itemize}

\section{II Related Work}
Traditional safety analysis relies on historical crash data and
regression models~\cite{gershon2019distracted}, which are inherently
reactive and cannot support real-time assessment. Surrogate safety
measures (SSMs) such as TTC~\cite{hayward1972near} and
PET~\cite{peesapati2018can} enable proactive conflict
detection~\cite{murat2017integration}, but typically depend on fixed
thresholds~\cite{paul2020post} and do not scale to multi-agent,
frame-level risk estimation. Machine learning methods --- including
SVMs, Random Forests, and deep learning --- improve predictive
performance~\cite{kucskapan2022pedestrian,nguyen2022assessment}, yet
most target specific user groups or rely on handcrafted features,
limiting their capacity to represent heterogeneous interactions.

Graph neural networks address this limitation by explicitly modeling
relational structure among traffic agents. Recent work has applied
heterogeneous GNNs to intersection safety, including multi-relational
GCNs for collision probability estimation~\cite{sonth2023real},
spatiotemporal GCNs with dynamic heterogeneous
graphs~\cite{peng2025pastgcn}, and graph-based frameworks with
distinct node types for pedestrians and
vehicles~\cite{muduli2026graph}. However, these efforts focus
exclusively on intersections and pedestrian scenarios. On freeways,
lane-changing and merging are the primary conflict contributors, and
combining TTC with PET improves crash estimation over either measure
alone~\cite{qi2020modified}, yet existing approaches analyze
individual conflict pairs in isolation rather than jointly modeling
the full multi-agent scene~\cite{xiang2024research}. Our work
addresses these gaps by unifying SSM-based conflict labeling with
heterogeneous graph modeling for frame-level freeway risk assessment.

\section{III Methodology}
\label{sec:methodology}

\subsection{Problem Formulation}
\label{subsec:problem_formulation}

This study considers \emph{real-time} traffic conflict risk prediction on freeway segments using high-resolution vehicle trajectory data. The raw data consist of kinematic observations recorded at discrete time steps (frames). The learning objective is to assign each frame a binary label---\emph{risky} or \emph{safe}---based on whether any surrogate safety measure (SSM) violation occurs among the vehicles observed in that frame.
Let $\mathcal{F}=\{f_1,f_2,\ldots,f_T\}$ denote the ordered set of frames. At time step $t$, frame $f_t$ contains a set of active vehicles $\mathcal{V}_t=\{v_1,v_2,\ldots,v_{N_t}\}$ with state attributes such as position, velocity, acceleration, lane assignment, and headway. We aim to learn a parameterized predictor $\hat{y}_t = g_\theta(\mathcal{V}_t, \mathcal{E}_t),$ which maps the vehicles and their interactions to a binary label $y_t\in\{0,1\}$, where $y_t=1$ indicates a risky frame and $y_t=0$ indicates a safe frame. Here, $\mathcal{E}_t$ represents the set of pairwise interaction relations among vehicles at time $t$. We cast the problem as \emph{graph-level} binary classification. Each frame is represented as a directed, relation-aware graph $G_t = (\mathcal{V}_t,\mathcal{E}_t,\mathbf{X}_t,\mathbf{E}_t),$ where $\mathbf{X}_t\in\mathbb{R}^{N_t\times d_0}$ is the node feature matrix and $\mathbf{E}_t$ denotes edge attributes encoding interaction dynamics. The ground-truth label $y_t$ is derived from SSMs computed directly from the trajectories.

\subsection{Surrogate Safety Measures for Risk Labeling}
\label{sec:ssm_labeling}

SSMs provide a quantitative framework for assessing traffic conflict severity without relying on observed crash records, which are inherently rare and spatially sparse on freeways. In this study, we employ two complementary SSMs that capture the two predominant conflict types on freeway facilities: rear-end conflicts arising from longitudinal closing interactions, and sideswipe conflicts arising from lateral lane-change maneuvers.

\subsubsection{Time-to-Collision (TTC)}
\label{subsubsec:ttc}

Time-to-Collision (TTC) measures the time remaining to a potential rear-end conflict between a same-lane follower $i$ and leader $j$. For actively closing pairs with a positive bumper-to-bumper gap, TTC is computed as $\text{TTC}_{ij}(t)=d_{ij}(t)/\Delta v_{ij}(t)$, where $\Delta v_{ij}(t)=v_i(t)-v_j(t)>v_{\min}$, $d_{ij}(t)$ is the space headway minus the leader length, and $v_{\min}=0.5$ ft/s excludes quasi-stationary pairs. TTC is evaluated only when $d_{ij}(t)>0$.

\subsubsection{Post-Encroachment Time (PET)}

\begin{figure}
    \centering
    \includegraphics[width=.7\linewidth]{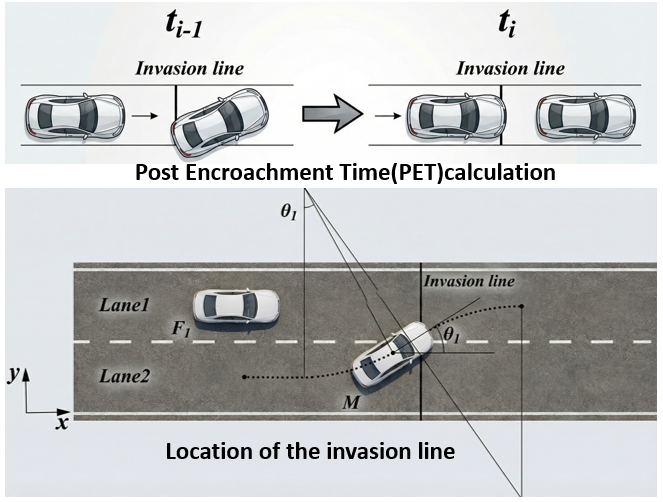}
    \caption{Illustration of PET computation on a freeway segment. \textbf{Top:} Temporal sequence showing the invasion line crossed by the preceding vehicle at time \(t_{i-1}\) and the lane-changing vehicle at time \(t_i\); PET is the elapsed time between these two occupancy events. \textbf{Bottom:} Spatial configuration of the conflict: vehicle \(F_t\) in Lane 1 and vehicle \(M\) performing a lane change into Lane 2, with the invasion line marking the shared road space where longitudinal overlap creates a potential sideswipe conflict point.}
    \label{fig:PET}
\end{figure}

Post-Encroachment Time captures the temporal margin between successive occupancies of a shared road space during lane-change maneuvers. Unlike TTC, which measures an ongoing longitudinal closing interaction, PET measures the temporal gap at a spatial conflict point that has already been traversed by one vehicle and subsequently entered by another (See Figure~\ref{fig:PET}).

We compute PET using a cell-based spatial occupancy method. The longitudinal extent of the freeway is discretized into cells of width $\Delta s$ (set to 5 ft). For each vehicle at each time step, the set of occupied cells is determined from the vehicle's front and rear positions. Upon detection of a sustained lane change — defined as a vehicle transitioning to an adjacent lane and maintaining the new lane assignment for at least $\tau_{\text{sustain}}$ consecutive frames (set to 1.0 s) — the algorithm identifies the most recent prior occupant of each cell in the target lane. The PET for the conflict pair is the minimum temporal gap across all shared cells $\text{PET}_{ij}(t) = \frac{t_{\text{entry},i} - t_{\text{exit},j}}{\text{FR}}$, where $t_{\text{entry},i}$ is the frame at which the lane-changing vehicle $i$ enters the target-lane cell, $t_{\text{exit},j}$ is the last frame at which the previously occupying vehicle $j$ was present in that cell, and FR is the frame rate (10 Hz). A minimum PET floor of 0.2 s is applied to exclude physically implausible detections arising from sensor noise.

\subsubsection{Threshold-Based Risk Labeling}
\label{sec:Threshold}

To evaluate model performance across varying levels of conflict severity, we define nine threshold configurations spanning three categories. For rear-end interactions, we consider TTC thresholds of $0.5\,\mathrm{s}$, $1.0\,\mathrm{s}$, and $1.5\,\mathrm{s}$, corresponding to imminent, critical, and precautionary conflict regimes, respectively. For lane-change interactions, we consider PET thresholds of $1.0\,\mathrm{s}$, $1.5\,\mathrm{s}$, and $2.0\,\mathrm{s}$ to represent close, moderate, and marginal sideswipe conflicts. Finally, we evaluate combined configurations using joint TTC$\mid$PET criteria---namely, $(\mathrm{TTC}<0.5\,\mathrm{s}\ \mid\ \mathrm{PET}<1.0\,\mathrm{s})$, $(\mathrm{TTC}<1.0\,\mathrm{s}\ \mid\ \mathrm{PET}<1.5\,\mathrm{s})$, and $(\mathrm{TTC}<1.5\,\mathrm{s}\ \mid\ \mathrm{PET}<2.0\,\mathrm{s})$---where a frame is labeled risky if \emph{either} condition is satisfied. This multi-threshold framework enables a systematic robustness analysis across heterogeneous conflict mechanisms and risk severities, ranging from highly imbalanced settings (e.g., $\mathrm{TTC}<0.5\,\mathrm{s}$ with approximately $5$--$14\%$ risk prevalence) to substantially more balanced label distributions (e.g., $\mathrm{TTC}<1.5\,\mathrm{s}\mid \mathrm{PET}<2.0\,\mathrm{s}$ with approximately $30$--$58\%$ prevalence).
For each frame $f_t$, the binary label is:
\begin{equation}
y_t = \begin{cases} 
1 & \text{if } \exists\, (i,j): \text{TTC}_{ij}(t) < \tau_{\text{TTC}} \;\text{or}\; \text{PET}_{ij}(t) < \tau_{\text{PET}} \\ 
0 & \text{otherwise} 
\end{cases}
\end{equation}
where $\tau_{\text{TTC}}$ and $\tau_{\text{PET}}$ are the respective thresholds for the configuration under evaluation.

\subsection{Graph Construction from Trajectory Data}

Each traffic frame is transformed into a heterogeneous graph that explicitly encodes the two distinct modes of vehicle interaction observed on freeways: longitudinal (same-lane) interactions and lateral (adjacent-lane) interactions.

\subsubsection{Node Representation}

Each vehicle $v_i$ present in frame $f_t$ is represented as a node with a $d$-dimensional feature vector $\mathbf{x}_i \in \mathbb{R}^{10}$ comprising:
\begin{equation}
\mathbf{x}_i = [\, x_i,\; y_i,\; v_i,\; a_i,\; \ell_i,\; s_i,\; h_i,\; L_i,\; \dot{x}_i,\; \lambda_i \,]
\end{equation}
where $x_i$ and $y_i$ are the lateral and longitudinal positions, $v_i$ is the longitudinal velocity, $a_i$ is the longitudinal acceleration, $\ell_i$ is the lane assignment, $s_i$ is the space headway to the preceding vehicle, $h_i$ is the time headway, $L_i$ is the vehicle length, $\dot{x}_i$ is the lateral velocity (computed as the finite difference of lateral position), and $\lambda_i$ is a lane-change indicator flag that equals 1 if the vehicle has changed lanes within the preceding 1.0 s window (10 frames). All features are standardized using robust z-score normalization with percentile-clipped statistics (1st and 99th percentiles) to mitigate the influence of outliers. For frames with more than 200 vehicles, a random subset of 200 nodes is sampled to maintain computational tractability.

\subsubsection{Edge Construction and Classification}

Edges are established between all vehicle pairs within a spatial proximity radius $r = 100$ ft. Each edge is classified into one of two types based on the lane relationship between the connected vehicles.  \textbf{Longitudinal edges} $\mathcal{E}^{\text{long}}$: connect vehicle pairs occupying the same lane ($|\ell_i - \ell_j| < 0.5$). These edges capture car-following dynamics relevant to rear-end conflict assessment. \textbf{Lateral edges} $\mathcal{E}^{\text{lat}}$: connect vehicle pairs in adjacent lanes ($0.5 < |\ell_i - \ell_j| < 1.5$). These edges encode cross-lane spatial relationships relevant to lane-change conflict assessment.

\subsubsection{Physics-Informed Edge Features}

Each edge type carries a distinct set of three physics-informed features designed to encode the interaction dynamics most relevant to its associated conflict type.
For longitudinal edges $(i, j) \in \mathcal{E}^{\text{long}}$, the edge feature vector is $\mathbf{e}^{\text{long}}_{ij} = \left[\, \frac{\Delta v_{ij}}{30},\;\; \frac{d_{ij}}{r},\;\; \frac{\Delta a_{ij}}{20} \,\right]$, where $\Delta v_{ij} = v_i - v_j$ is the closing rate (directly related to TTC), $d_{ij}$ is the Euclidean distance normalized by the edge radius, and $\Delta a_{ij} = a_i - a_j$ is the acceleration differential indicating whether the closing rate is increasing or decreasing. For lateral edges $(i, j) \in \mathcal{E}^{\text{lat}}$, the edge feature vector is $\mathbf{e}^{\text{lat}}_{ij} = \left[\, \frac{\dot{x}_i}{5},\;\; \lambda_i,\;\; \frac{o_{ij}}{15} \,\right]$, where $\dot{x}_i$ is the lateral velocity of the source vehicle, $\lambda_i$ is its lane-change flag, and $o_{ij}$ is the longitudinal overlap between the two vehicles computed as $o_{ij} = \max(0, \min(y_i^{\text{front}}, y_j^{\text{front}}) - \max(y_i^{\text{rear}}, y_j^{\text{rear}}))$. The overlap feature captures the spatial proximity condition that makes sideswipe conflicts physically possible — a lane change is only dangerous if the vehicles share longitudinal road space.

This edge-type separation is a deliberate design choice. The closing rate $\Delta v_{ij}$ is the critical feature for predicting TTC-based conflicts, while the lateral velocity and overlap are critical for predicting PET-based conflicts. By providing each interaction type through its own dedicated edge channel, the model receives physics-aligned information that reflects the distinct mechanisms underlying rear-end and sideswipe conflicts.

\subsection{Proposed Architecture: HIA-GAT}

\begin{figure*}[htbp]
    \centering
    \includegraphics[width=1\linewidth]{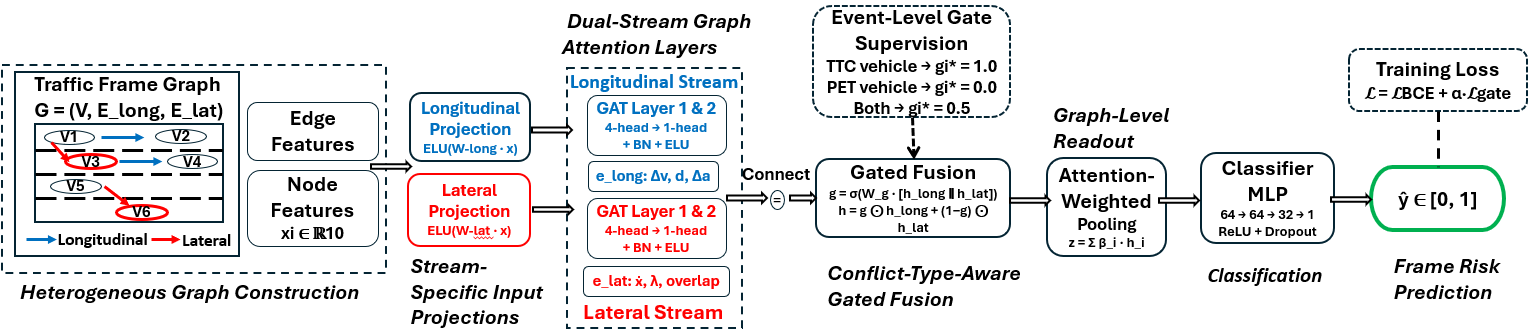}
    \caption{Overall architecture of the proposed HIA-GAT model}
    \label{fig:architecture}
\end{figure*}

We propose the Heterogeneous Interaction-Aware Graph Attention Network (HIA-GAT), a dual-stream graph neural network architecture that processes longitudinal and lateral vehicle interactions through dedicated pathways and fuses them via a learned, conflict-type-aware gating mechanism. Figure~\ref{fig:architecture} presents the overall architecture.

\subsubsection{Stream-Specific Input Projections}

To enable each stream to develop specialized representations, the shared node features $\mathbf{x}_i$ are first transformed through stream-specific input projections $\mathbf{x}_i^{\text{long}} = \text{ELU}(\mathbf{W}^{\text{long}} \mathbf{x}_i + \mathbf{b}^{\text{long}}), \quad
\mathbf{x}_i^{\text{lat}} = \text{ELU}(\mathbf{W}^{\text{lat}} \mathbf{x}_i + \mathbf{b}^{\text{lat}})$, where $\mathbf{W}^{\text{long}}, \mathbf{W}^{\text{lat}} \in \mathbb{R}^{d \times d}$ are learnable weight matrices. These projections allow the longitudinal stream to emphasize features relevant to car-following dynamics (e.g., velocity, headway) while the lateral stream can attend more to lane-change indicators (e.g., lateral velocity, lane-change flag).

\subsubsection{Dual-Stream Graph Attention Layers}

Each stream employs a two-layer Graph Attention Network (GAT) that operates exclusively on its corresponding edge type. The longitudinal stream processes messages along same-lane edges $\mathcal{E}^{\text{long}}$ with longitudinal edge features $\mathbf{e}^{\text{long}}_{ij}$, while the lateral stream processes messages along adjacent-lane edges $\mathcal{E}^{\text{lat}}$ with lateral edge features $\mathbf{e}^{\text{lat}}_{ij}$. For the longitudinal stream, the first GAT layer with multi-head attention computes $\alpha_{ij}^{k} = \frac{\exp\left(\text{LeakyReLU}\left(\mathbf{a}^{k\top}[\mathbf{W}^{k}\mathbf{x}_i^{\text{long}} \| \mathbf{W}^{k}\mathbf{x}_j^{\text{long}} \| \mathbf{W}^{k}_{e}\mathbf{e}_{ij}^{\text{long}}]\right)\right)}{\sum_{m \in \mathcal{N}^{\text{long}}_i} \exp\left(\text{LeakyReLU}\left(\mathbf{a}^{k\top}[\mathbf{W}^{k}\mathbf{x}_i^{\text{long}} \| \mathbf{W}^{k}\mathbf{x}_m^{\text{long}} \| \mathbf{W}^{k}_{e}\mathbf{e}_{im}^{\text{long}}]\right)\right)}$, and  
$\mathbf{h}_i^{\text{long},(1)} = \text{ELU}\left(\text{BN}\left( \Big\|_{k=1}^{K} \sum_{j \in \mathcal{N}^{\text{long}}_i} \alpha_{ij}^{k} \mathbf{W}^{k} \mathbf{x}_j^{\text{long}} \right)\right)$.
where $K$ is the number of attention heads (set to 4), $\|$ denotes concatenation, $\mathcal{N}^{\text{long}}_i$ is the set of same-lane neighbors of vehicle $i$, and BN denotes batch normalization. The edge features $\mathbf{e}_{ij}^{\text{long}}$ are incorporated into the attention coefficient computation, enabling the model to modulate message strength based on the physical interaction dynamics (e.g., assigning higher attention to vehicle pairs with large closing rates).

The second GAT layer uses single-head attention to produce the final per-node longitudinal embedding $\mathbf{h}_i^{\text{long}} \in \mathbb{R}^{H}$, where $H$ is the hidden dimension (set to 64). The lateral stream follows an identical architecture but operates on lateral edges and edge features, producing $\mathbf{h}_i^{\text{lat}} \in \mathbb{R}^{H}$.

\subsubsection{Conflict-Type-Aware Gated Fusion}

The central architectural innovation of HIA-GAT is a per-node gating mechanism that dynamically determines the relative contribution of the longitudinal and lateral streams based on each vehicle's involvement in different conflict types. For each node $i$, the gate value is computed from the concatenated stream embeddings $\mathbf{g}_i = \sigma\left(\mathbf{W}_g [\mathbf{h}_i^{\text{long}} \| \mathbf{h}_i^{\text{lat}}] + \mathbf{b}_g\right) \in (0, 1)^{H}$, where $\sigma$ is the sigmoid activation and $\mathbf{W}_g \in \mathbb{R}^{H \times 2H}$ is a learnable weight matrix. The fused node embedding is then
$\mathbf{h}_i = \mathbf{g}_i \odot \mathbf{h}_i^{\text{long}} + (1 - \mathbf{g}_i) \odot \mathbf{h}_i^{\text{lat}}$, 
where $\odot$ denotes element-wise multiplication. A gate value close to 1.0 routes the representation through the longitudinal stream, while a value close to 0.0 routes it through the lateral stream. Vehicles involved in both conflict types receive intermediate gate values.

\subsubsection{Event-Level Gate Supervision}

A key challenge in training gated fusion architectures is ensuring that the gate learns to differentiate between conflict types rather than collapsing to a trivial solution. We address this through an event-level supervision mechanism that leverages the conflict-type information inherently available from the SSM computation.

During the risk labeling process (Section~\ref{sec:ssm_labeling}), we identify not only the frame-level risk label but also the specific vehicles participating in each detected conflict event. We leverage this event attribution to define per-node gate supervision targets $g_i^{*}$, assigning $g_i^{*}=1.0$ when vehicle $i$ participates in a TTC-based conflict pair, $g_i^{*}=0.0$ when it participates in a PET-based conflict pair, and $g_i^{*}=0.5$ when it is simultaneously implicated in both TTC and PET conflicts; conversely, if vehicle $i$ is not involved in any conflict event, its gate target is masked and excluded from supervision.

The gate supervision loss is computed as the mean squared error between the predicted gate values and the targets, applied only to supervised nodes $\mathcal{L}_{\text{gate}} = \frac{1}{|\mathcal{M}|} \sum_{i \in \mathcal{M}} \| \bar{\mathbf{g}}_i - g_i^{*} \|^2$, where $\mathcal{M}$ is the set of supervised nodes (those involved in at least one conflict event) and $\bar{\mathbf{g}}_i$ is the mean gate value across hidden dimensions for node $i$. Critically, only a small fraction of nodes receive supervision — typically 0.1–1.5\% of all nodes in a frame — yet the gate generalizes to unsupervised nodes through the shared gate projection layer, which learns to map node representations to conflict-type likelihood.

\subsubsection{Graph-Level Readout and Classification}

The fused per-node embeddings are aggregated into a single graph-level representation using attention-weighted global pooling: $\mathbf{z}_t = \sum_{i \in \mathcal{V}_t} \beta_i \, \mathbf{h}_i, \quad \text{where} \quad \beta_i = \frac{\exp\left(\text{MLP}_{\text{att}}(\mathbf{h}_i)\right)}{\sum_{j \in \mathcal{V}_t} \exp\left(\text{MLP}_{\text{att}}(\mathbf{h}_j)\right)}$, where $\text{MLP}_{\text{att}}$ is a two-layer neural network that computes the attention score for each node. This attention pooling mechanism allows the model to focus on the most informative vehicles (typically those involved in conflict events) when producing the graph-level representation, rather than treating all vehicles equally.

The graph embedding $\mathbf{z}_t \in \mathbb{R}^{H}$ is passed through a three-layer MLP classifier $\hat{y}_t = \text{MLP}_{\text{cls}}(\mathbf{z}_t) = \mathbf{W}_3 \cdot \text{ReLU}(\mathbf{W}_2 \cdot \text{ReLU}(\mathbf{W}_1 \mathbf{z}_t))$, with hidden dimensions $H \to H \to H/2 \to 1$ and dropout regularization between layers. The output logit is transformed to a probability via the sigmoid function.

\subsubsection{Training Objective}

The total training loss combines the frame-level binary cross-entropy classification loss with the gate supervision loss $\mathcal{L} = \mathcal{L}_{\text{BCE}} + \alpha \, \mathcal{L}_{\text{gate}}$, 
where $\mathcal{L}_{\text{BCE}} = -\frac{1}{N}\sum_{t}\left[y_t \log(\hat{p}_t) + (1-y_t)\log(1-\hat{p}_t)\right]$ is the weighted binary cross-entropy loss with inverse class-frequency weighting to address class imbalance, and $\alpha$ is the gate supervision weight (set to 0.5). The model is trained using the Adam optimizer with a learning rate of $3 \times 10^{-4}$, weight decay of $10^{-4}$, gradient clipping at norm 1.0, learning rate warmup over the first 5 epochs, and ReduceLROnPlateau scheduling. Early stopping with a patience of 15 epochs is applied based on validation loss.

\subsection{Baseline Methods}

To comprehensively evaluate the proposed HIA-GAT, we compare against seven baseline methods organized in three tiers of increasing architectural complexity:

\textbf{Non-graph baselines} operate on frame-level aggregate feature vectors (70 dimensions) constructed by computing statistical summaries (mean, standard deviation, minimum, maximum) of the 10 node features across all vehicles in the frame, supplemented with graph structural statistics (node count, edge type ratios) and edge feature statistics (mean, standard deviation, minimum, maximum of the 3-dimensional longitudinal and lateral edge features). These baselines include Logistic Regression, Random Forest (200 trees), XGBoost (200 estimators), and a multi-layer perceptron (MLP) with two hidden layers.

\textbf{GNN baselines without edge features} include GCN and GraphSAGE, which operate on the homogeneous graph (all edges combined) using only node features. These baselines quantify the contribution of graph topology alone, independent of physics-informed edge attributes.

\textbf{Single-stream GNN baseline (HomoGAT)} uses the same GAT architecture as each individual stream in HIA-GAT, operating on the combined homogeneous edge set with 3-dimensional edge features (closing rate, distance, lane difference). This baseline isolates the contribution of the dual-stream decomposition and gated fusion mechanism by providing edge features to a single unified attention network.

All GNN models share identical hyperparameters (hidden dimension 64, 4 attention heads, dropout 0.3, learning rate $3 \times 10^{-4}$, patience 15) and use the same data splits (70/15/15 train/validation/test with stratified sampling and fixed random seed) to ensure fair comparison. Performance is evaluated using both threshold-optimized F1 score (F1$_\text{opt}$, where the classification threshold is selected to maximize F1 on the validation set) and Area Under the ROC Curve (AUC), which provides a threshold-independent measure of discriminative ability.

\section{IV Experiments}

\subsection{Dataset and Data Labeling}

We use NGSIM vehicle trajectories~\cite{ngsim_i80_2016,ngsim_us101_2016} recorded at 10\,Hz on two California freeway segments: I-80 in Emeryville, with 45 minutes of PM peak congestion, and US-101 in Los Angeles, with 45 minutes of AM peak traffic in a weaving section. Each record includes vehicle position, velocity, acceleration, lane, dimensions, and headway. Preprocessing disambiguated overlapping Frame\_ID sequences, removed duplicate boundary records, and added two derived features: lateral velocity and a lane-change flag indicating a sustained lane change within the previous 1.0\,s. The final I-80 and US-101 datasets contain 4,564,923 and 4,098,933 vehicle-frame records over 29,679 and 28,156 unique frames, respectively. Frame-level risk labels are generated for the nine threshold configurations in Section~\ref{sec:Threshold}. Experiments were run on an NVIDIA A100-SXM4-80GB GPU using PyTorch Geometric.

\begin{table}[htbp]
\centering
\footnotesize
\setlength{\tabcolsep}{3pt}
\caption{Performance comparison on the I-80 and US-101 datasets. Best AUC per configuration is bolded. \%R denotes risk prevalence. Each cell reports F1 score (left) and AUC (right). Shaded rows summarize category averages.}
\label{tab:performance}
\resizebox{\textwidth}{!}{
\begin{tabular}{l*{8}{c}}
\toprule
Configuration (\%R) & LogReg & RF & XGB & MLP & GCN & SAGE & HomoGAT & HIA-GAT \\
\midrule
\multicolumn{9}{c}{\textbf{I-80 Dataset}} \\
TTC$<$0.5 (13.9)   & 0.383 / 0.755 & 0.547 / 0.877 & 0.538 / \textbf{0.879} & 0.469 / 0.833 & 0.390 / 0.763 & 0.423 / 0.796 & 0.434 / 0.802 & 0.449 / 0.829 \\
TTC$<$1.0 (33.5)   & 0.624 / 0.780 & 0.767 / \textbf{0.902} & 0.753 / 0.899 & 0.694 / 0.854 & 0.628 / 0.799 & 0.641 / 0.807 & 0.668 / 0.826 & 0.730 / 0.886 \\
TTC$<$1.5 (55.2)   & 0.795 / 0.824 & 0.868 / \textbf{0.924} & 0.861 / 0.916 & 0.828 / 0.880 & 0.789 / 0.828 & 0.796 / 0.838 & 0.811 / 0.853 & 0.835 / 0.889 \\
\rowcolor{gray!15}
\textit{TTC Avg}    & \textit{0.601 / 0.786} & \textit{0.727 / \textbf{0.901}} & \textit{0.717 / 0.898} & \textit{0.664 / 0.856} & \textit{0.602 / 0.797} & \textit{0.620 / 0.814} & \textit{0.638 / 0.827} & \textit{0.671 / 0.868} \\
PET$<$1.0 (4.6)    & 0.174 / 0.730 & 0.145 / 0.703 & 0.132 / 0.684 & 0.166 / 0.721 & 0.195 / 0.758 & 0.209 / \textbf{0.816} & 0.231 / 0.797 & 0.217 / 0.808 \\
PET$<$1.5 (5.9)    & 0.226 / 0.748 & 0.193 / 0.725 & 0.175 / 0.694 & 0.221 / 0.744 & 0.214 / 0.763 & 0.241 / \textbf{0.811} & 0.273 / 0.806 & 0.245 / 0.805 \\
PET$<$2.0 (6.3)    & 0.218 / 0.745 & 0.194 / 0.717 & 0.171 / 0.690 & 0.206 / 0.739 & 0.229 / 0.770 & 0.266 / 0.804 & 0.268 / \textbf{0.811} & 0.265 / 0.803 \\
\rowcolor{gray!15}
\textit{PET Avg}    & \textit{0.206 / 0.741} & \textit{0.177 / 0.715} & \textit{0.159 / 0.689} & \textit{0.198 / 0.735} & \textit{0.213 / 0.764} & \textit{0.239 / \textbf{0.810}} & \textit{0.257 / 0.805} & \textit{0.242 / 0.805} \\
TTC$<$0.5$|$PET$<$1.0 (17.7) & 0.409 / 0.732 & 0.519 / \textbf{0.812} & 0.515 / 0.812 & 0.456 / 0.760 & 0.411 / 0.727 & 0.417 / 0.734 & 0.435 / 0.760 & 0.453 / 0.775 \\
TTC$<$1.0$|$PET$<$1.5 (37.1) & 0.620 / 0.754 & 0.739 / \textbf{0.864} & 0.731 / 0.855 & 0.673 / 0.809 & 0.639 / 0.772 & 0.646 / 0.776 & 0.664 / 0.797 & 0.702 / 0.840 \\
TTC$<$1.5$|$PET$<$2.0 (57.5) & 0.796 / 0.818 & 0.860 / \textbf{0.912} & 0.852 / 0.904 & 0.826 / 0.871 & 0.797 / 0.822 & 0.797 / 0.827 & 0.803 / 0.839 & 0.837 / 0.884 \\
\midrule
Average             & 0.472 / 0.765 & 0.537 / 0.826 & 0.525 / 0.815 & 0.504 / 0.801 & 0.477 / 0.778 & 0.493 / 0.801 & 0.509 / 0.810 & 0.525 / \textbf{0.835} \\
\midrule
\multicolumn{9}{c}{\textbf{US-101 Dataset}} \\
TTC$<$0.5 (5.6)   & 0.232 / 0.809 & 0.472 / 0.928 & 0.547 / \textbf{0.952} & 0.428 / 0.914 & 0.288 / 0.814 & 0.281 / 0.822 & 0.373 / 0.874 & 0.248 / 0.774 \\
TTC$<$1.0 (13.7)  & 0.428 / 0.793 & 0.690 / 0.944 & 0.713 / \textbf{0.949} & 0.650 / 0.933 & 0.469 / 0.834 & 0.472 / 0.846 & 0.548 / 0.884 & 0.566 / 0.902 \\
TTC$<$1.5 (27.4)  & 0.608 / 0.813 & 0.807 / \textbf{0.947} & 0.793 / 0.940 & 0.746 / 0.923 & 0.637 / 0.840 & 0.655 / 0.863 & 0.650 / 0.861 & 0.706 / 0.899 \\
\rowcolor{gray!15}
\textit{TTC Avg}    & \textit{0.423 / 0.805} & \textit{0.656 / \textbf{0.940}} & \textit{0.684 / 0.947} & \textit{0.608 / 0.923} & \textit{0.465 / 0.829} & \textit{0.469 / 0.844} & \textit{0.524 / 0.873} & \textit{0.507 / 0.858} \\
PET$<$1.0 (3.5)   & 0.166 / 0.757 & 0.151 / 0.728 & 0.131 / 0.687 & 0.154 / 0.758 & 0.168 / 0.777 & 0.235 / 0.859 & 0.225 / 0.834 & 0.256 / \textbf{0.875} \\
PET$<$1.5 (4.2)   & 0.194 / 0.768 & 0.144 / 0.735 & 0.149 / 0.718 & 0.190 / 0.768 & 0.161 / 0.774 & 0.254 / \textbf{0.880} & 0.228 / 0.850 & 0.276 / 0.870 \\
PET$<$2.0 (4.4)   & 0.212 / 0.790 & 0.197 / 0.773 & 0.177 / 0.753 & 0.221 / 0.779 & 0.202 / 0.769 & 0.267 / 0.855 & 0.295 / \textbf{0.874} & 0.285 / 0.873 \\
\rowcolor{gray!15}
\textit{PET Avg}    & \textit{0.191 / 0.772} & \textit{0.164 / 0.745} & \textit{0.152 / 0.719} & \textit{0.188 / 0.768} & \textit{0.177 / 0.773} & \textit{0.252 / 0.865} & \textit{0.249 / 0.853} & \textit{0.272 / \textbf{0.873}} \\
TTC$<$0.5$|$PET$<$1.0 (8.9) & 0.296 / 0.771 & 0.457 / \textbf{0.849} & 0.423 / 0.831 & 0.352 / 0.798 & 0.275 / 0.735 & 0.372 / 0.841 & 0.347 / 0.805 & 0.385 / 0.840 \\
TTC$<$1.0$|$PET$<$1.5 (17.3) & 0.493 / 0.798 & 0.666 / \textbf{0.900} & 0.658 / 0.897 & 0.576 / 0.870 & 0.509 / 0.812 & 0.526 / 0.837 & 0.558 / 0.863 & 0.609 / 0.888 \\
TTC$<$1.5$|$PET$<$2.0 (30.3) & 0.614 / 0.800 & 0.787 / \textbf{0.918} & 0.762 / 0.910 & 0.724 / 0.891 & 0.635 / 0.811 & 0.652 / 0.830 & 0.676 / 0.859 & 0.710 / 0.881 \\
\midrule
Average             & 0.360 / 0.789 & 0.486 / 0.858 & 0.484 / 0.849 & 0.449 / 0.848 & 0.372 / 0.796 & 0.413 / 0.848 & 0.433 / 0.856 & 0.449 / \textbf{0.867} \\
\bottomrule
\end{tabular}%
}
\end{table}


\subsection{Evaluation Metrics}
We evaluate all models using two complementary metrics. The primary metric is the Area Under the Receiver Operating Characteristic Curve (AUC), which measures the model's ability to rank frames by risk severity independently of any classification threshold. AUC is particularly suited to traffic conflict prediction because the operational goal is risk prioritization — identifying which frames warrant attention — rather than binary classification at a fixed threshold. Moreover, AUC is robust to class imbalance, which varies substantially across our nine configurations (from 3.5\% to 57.5\% risk prevalence).
As a secondary metric, we report the threshold-optimized F1 score, where the classification threshold is selected to maximize F1 on the validation set. F1 provides a measure of the best achievable classification performance but is sensitive to the probability distribution shape and class balance, making it less stable for cross-configuration comparison.

\subsection{Overall Comparison}
Table~\ref{tab:performance} present the full comparison of all eight methods across the nine threshold configurations on the I-80 and US-101 datasets, respectively. Two clear patterns emerge from the aggregate results. First, the proposed HIA-GAT achieves the highest average AUC on both datasets (0.835 on I-80, 0.867 on US-101), outperforming all non-graph and graph baselines. Second, Random Forest achieves the highest average F1 (0.537 on I-80, 0.486 on US-101). This divergence between AUC and F1 rankings reflects a fundamental distinction: AUC evaluates the full ranking of risk scores across all operating points, whereas F1 evaluates classification accuracy at a single best threshold. HIA-GAT produces well-calibrated probability scores that capture nuanced risk gradations, while tree-based ensembles produce sharper probability distributions that yield cleaner decision boundaries at optimal thresholds despite slightly inferior overall ranking.
Among the GNN baselines, a consistent architectural progression is observed. GCN, which operates on graph topology alone without edge features, achieves the lowest GNN performance (average AUC of 0.778 on I-80, 0.796 on US-101). GraphSAGE improves upon GCN through its neighborhood sampling mechanism but similarly lacks edge feature integration. HomoGAT, which incorporates 3-dimensional physics-informed edge features into the attention computation, yields a substantial improvement over GCN (+0.032 AUC on I-80, +0.060 on US-101), demonstrating that edge features encoding pairwise vehicle dynamics are critical for risk prediction. The proposed HIA-GAT extends this further with its dual-stream architecture and event-level gate supervision, achieving +0.025 AUC over HomoGAT on I-80 and +0.011 on US-101, and winning 7 of 9 configurations on both datasets by AUC.
\subsection{Conflict-Type Stratified Analysis}
The nine threshold configurations can be partitioned into three categories — TTC-only (rear-end conflicts), PET-only (lane-change conflicts), and combined — revealing distinct patterns in model suitability.
On TTC-only configurations, non-graph models perform strongly due to the nature of rear-end conflict detection. TTC is fundamentally determined by the closing rate and gap distance between a following vehicle and its leader — quantities that are well captured by aggregate frame-level statistics such as the mean and maximum closing rate across all vehicle pairs. Random Forest achieves average TTC AUC of 0.901 on I-80 and 0.940 on US-101. GNN models remain competitive (HIA-GAT: 0.868 and 0.858) but the additional representational capacity of graph-level reasoning provides diminishing returns when the risk signal is already well-encoded in simple summary statistics.
On PET-only configurations, the pattern reverses decisively. Non-graph methods degrade substantially, with RF and XGBoost achieving average PET AUC of only 0.715 and 0.689 on I-80, respectively (Table \ref{tab:performance}). In contrast, all GNN models outperform all non-graph baselines, with HIA-GAT and GraphSAGE achieving 0.805 and 0.810 on I-80, and 0.873 and 0.865 on US-101. This gap — approximately 0.10 AUC separating graph from non-graph methods — provides strong evidence that graph structure is essential for lane-change conflict detection. PET conflicts are inherently relational events: they arise from the spatial interaction between a lane-changing vehicle and an adjacent-lane occupant sharing longitudinal road space. Aggregate statistics collapse this pairwise structure, losing the specific vehicle pair information that makes PET detection possible.
On combined configurations (TTC|PET), the results reflect an intermediate regime. RF leads overall due to its strong TTC performance, but HIA-GAT achieves the largest improvement over HomoGAT in this category (+0.034 AUC on I-80, +0.028 on US-101), suggesting that the dual-stream gating mechanism provides its greatest value when both conflict types coexist within the same frame and must be differentiated. The exception is $\mathrm{TTC}<0.5$ on US-101 (5.6\% prevalence), where HIA-GAT underperforms on both F1 and AUC due to training instability on this extremely sparse configuration --- the dual-stream architecture splits an already limited set of positive examples across two streams, reducing the effective supervision signal per stream below the threshold needed for stable attention learning.

\subsection{Gate Supervision Analysis}
A central design feature of HIA-GAT is the conflict-type aware gating mechanism. Table~\ref{tab:gate} summarizes the gate behavior across all nine configurations on both datasets. We evaluate whether the gate successfully learns to differentiate between longitudinal and lateral conflicts through its supervision signal. On single-type configurations, the gate achieves near-perfect conflict type routing. For TTC-only configurations, the average gate value across all nodes converges above 0.5 (0.69-0.77 on I-80, 0.58--0.67 on US-101), indicating that the majority of the representation is routed through the longitudinal stream. For PET-only configurations, the gate converges below 0.5 (0.43--0.46 on I-80, 0.40--0.43 on US-101), routing the representation predominantly through the lateral stream. The gate direction accuracy -- defined as the fraction of supervised nodes where the gate value correctly exceeds 0.5 for TTC-involved vehicles or falls below 0.5 for PET-involved vehicles --- exceeds 99\% on all single-type configurations across both datasets.
\begin{figure}
    \centering
    \includegraphics[width=1\linewidth]{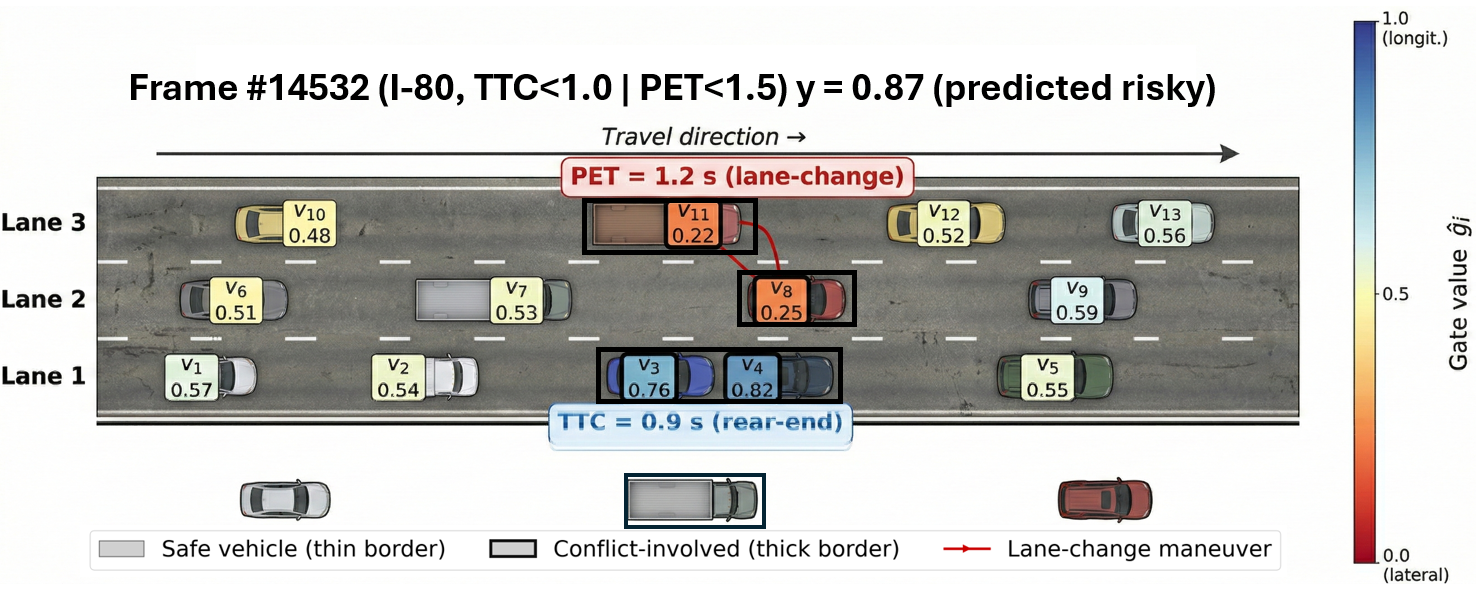}
    \caption{Per-vehicle gate values on a representative I-80 test frame (TTC$<$1.0\,|\,PET$<$1.5). Blue vehicles ($\bar{g}_i \to 1.0$) are routed through the longitudinal stream; red vehicles ($\bar{g}_i \to 0.0$) through the lateral stream. The gate correctly identifies the rear-end conflict pair ($v_3$/$v_4$) and the lane-change conflict pair ($v_8$/$v_{11}$), while safe vehicles remain near 0.5.}
    \label{fig:gate}
\end{figure}
On combined configurations, the gate exhibits substantial per-node variation (standard deviation 0.19--0.29), indicating that different vehicles within the same frame are routed through different streams depending on their conflict involvement. The aggregate gate mean shifts toward the TTC-dominant direction (0.54--0.70) because TTC events are approximately 6--11 times more numerous than PET events, which reduces PET gate direction accuracy on the most imbalanced combined configurations. Notably, only 0.1--1.5\% of nodes in each frame receive gate supervision, yet the gate generalizes to all nodes through the shared projection layer --- demonstrating that the model learns a mapping from node representations to conflict-type likelihood rather than memorizing supervised examples.

\begin{table}[t]
\centering
\footnotesize
\setlength{\tabcolsep}{2.5pt}
\caption{Gate supervision analysis for HIA-GAT. $\bar{g}$: mean gate value across all nodes (1.0 = fully longitudinal, 0.0 = fully lateral). $\sigma_g$: standard deviation. Dir.\ Acc.: fraction of supervised nodes with correct gate direction ($>$0.5 for TTC, $<$0.5 for PET).}
\label{tab:gate}
\begin{tabular}{l cc c cc c}
\toprule
 & \multicolumn{3}{c}{\textbf{I-80}} & \multicolumn{3}{c}{\textbf{US-101}} \\
\cmidrule(lr){2-4} \cmidrule(lr){5-7}
Configuration & $\bar{g}$ & $\sigma_g$ & Dir.\ Acc. & $\bar{g}$ & $\sigma_g$ & Dir.\ Acc. \\
\midrule
\multicolumn{7}{l}{\textit{TTC-only (expected: $\bar{g} > 0.5$)}} \\
TTC$<$0.5   & 0.71 & 0.08 & 100\%  & 0.67 & 0.23 & 99.7\% \\
TTC$<$1.0   & 0.69 & 0.09 & 100\%  & 0.58 & 0.24 & 99.9\% \\
TTC$<$1.5   & 0.77 & 0.07 & 100\%  & 0.60 & 0.25 & 100\%  \\
\midrule
\multicolumn{7}{l}{\textit{PET-only (expected: $\bar{g} < 0.5$)}} \\
PET$<$1.0   & 0.46 & 0.12 & 99.9\% & 0.40 & 0.12 & 99.1\% \\
PET$<$1.5   & 0.46 & 0.13 & 99.9\% & 0.41 & 0.15 & 100\%  \\
PET$<$2.0   & 0.43 & 0.14 & 100\%  & 0.43 & 0.15 & 100\%  \\
\midrule
\multicolumn{7}{l}{\textit{Combined (expected: per-node variation)}} \\
TTC$<$0.5$|$PET$<$1.0 & 0.60 & 0.23 & 89--91\% & 0.54 & 0.25 & 69--77\% \\
TTC$<$1.0$|$PET$<$1.5 & 0.64 & 0.25 & 88--93\% & 0.55 & 0.27 & 77--99\% \\
TTC$<$1.5$|$PET$<$2.0 & 0.70 & 0.19 & 80--95\% & 0.56 & 0.29 & 18--99\% \\
\bottomrule
\end{tabular}
\end{table}

\subsection{Discussion}
The most consequential finding of this study is the stark performance gap between graph-based and non-graph methods on PET-only configurations. While non-graph baselines achieve competitive or superior performance on TTC detection, they consistently fail on PET detection, with AUC values in the 0.69–0.77 range compared to 0.80–0.87 for GNN models. This finding has a clear physical interpretation. Rear-end conflicts are characterized by aggregate kinematic signatures — high mean closing rates, small minimum headways — that are preserved by statistical summaries of frame-level features. Lane-change conflicts, by contrast, are inherently pairwise events defined by the spatial and temporal relationship between two specific vehicles: one changing lanes and one occupying the target space. When these pairwise interactions are collapsed into frame-level means and maxima, the critical relational information is lost.
This result suggests that graph-based representations are not merely a modeling convenience for traffic risk prediction, but a necessary structural choice for comprehensive conflict detection that spans both longitudinal and lateral interaction types.

The comparison between GCN/GraphSAGE and HomoGAT/HIA-GAT isolates the value of physics-informed edge features. On I-80 TTC configurations, HomoGAT improves over GCN by $+0.030$ average AUC by explicitly encoding closing rate, gap distance, and acceleration differential—the key variables underlying TTC. HIA-GAT further extends this idea through conflict-specific streams: the longitudinal stream uses closing rate and acceleration differential, while the lateral stream uses lateral velocity, lane-change flag, and longitudinal overlap. This prevents one attention function from learning two distinct conflict mechanisms simultaneously. The gate also improves interpretability by indicating the dominant conflict type for each risky vehicle: values near 1.0 correspond to TTC-related rear-end interactions, values near 0.0 to PET-related lane-change conflicts, and safe vehicles remain near 0.5, as shown in Figure~\ref{fig:gate}. This attribution is obtained from SSM-derived labels with negligible added annotation or computational cost.

\subsection{Dataset and Data Labeling}

We use NGSIM vehicle trajectories~\cite{ngsim_i80_2016,ngsim_us101_2016} recorded at 10\,Hz on two California freeway segments: I-80 in Emeryville, with 45 minutes of PM peak congestion, and US-101 in Los Angeles, with 45 minutes of AM peak traffic in a weaving section. Each record includes vehicle position, velocity, acceleration, lane, dimensions, and headway. Preprocessing disambiguated overlapping Frame\_ID sequences, removed duplicate boundary records, and added two derived features: lateral velocity and a lane-change flag indicating a sustained lane change within the previous 1.0\,s. The final I-80 and US-101 datasets contain 4,564,923 and 4,098,933 vehicle-frame records over 29,679 and 28,156 unique frames, respectively. Frame-level risk labels are generated for the nine threshold configurations in Section~\ref{sec:Threshold}. Experiments were run on an NVIDIA A100-SXM4-80GB GPU using PyTorch Geometric.

\begin{table*}[htbp]
\centering
\footnotesize
\setlength{\tabcolsep}{3pt}
\caption{Performance comparison on the I-80 and US-101 datasets. Best AUC per configuration is bolded. \%R denotes risk prevalence. Each cell reports F1 score (left) and AUC (right). Shaded rows summarize category averages.}
\label{tab:performance}
\begin{tabular}{l*{8}{c}}
\toprule
Configuration (\%R) & LogReg & RF & XGB & MLP & GCN & SAGE & HomoGAT & HIA-GAT \\
\midrule
\multicolumn{9}{c}{\textbf{I-80 Dataset}} \\
TTC$<$0.5 (13.9)   & 0.383 / 0.755 & 0.547 / 0.877 & 0.538 / \textbf{0.879} & 0.469 / 0.833 & 0.390 / 0.763 & 0.423 / 0.796 & 0.434 / 0.802 & 0.449 / 0.829 \\
TTC$<$1.0 (33.5)   & 0.624 / 0.780 & 0.767 / \textbf{0.902} & 0.753 / 0.899 & 0.694 / 0.854 & 0.628 / 0.799 & 0.641 / 0.807 & 0.668 / 0.826 & 0.730 / 0.886 \\
TTC$<$1.5 (55.2)   & 0.795 / 0.824 & 0.868 / \textbf{0.924} & 0.861 / 0.916 & 0.828 / 0.880 & 0.789 / 0.828 & 0.796 / 0.838 & 0.811 / 0.853 & 0.835 / 0.889 \\
\rowcolor{gray!15}
\textit{TTC Avg}    & \textit{0.601 / 0.786} & \textit{0.727 / \textbf{0.901}} & \textit{0.717 / 0.898} & \textit{0.664 / 0.856} & \textit{0.602 / 0.797} & \textit{0.620 / 0.814} & \textit{0.638 / 0.827} & \textit{0.671 / 0.868} \\
PET$<$1.0 (4.6)    & 0.174 / 0.730 & 0.145 / 0.703 & 0.132 / 0.684 & 0.166 / 0.721 & 0.195 / 0.758 & 0.209 / \textbf{0.816} & 0.231 / 0.797 & 0.217 / 0.808 \\
PET$<$1.5 (5.9)    & 0.226 / 0.748 & 0.193 / 0.725 & 0.175 / 0.694 & 0.221 / 0.744 & 0.214 / 0.763 & 0.241 / \textbf{0.811} & 0.273 / 0.806 & 0.245 / 0.805 \\
PET$<$2.0 (6.3)    & 0.218 / 0.745 & 0.194 / 0.717 & 0.171 / 0.690 & 0.206 / 0.739 & 0.229 / 0.770 & 0.266 / 0.804 & 0.268 / \textbf{0.811} & 0.265 / 0.803 \\
\rowcolor{gray!15}
\textit{PET Avg}    & \textit{0.206 / 0.741} & \textit{0.177 / 0.715} & \textit{0.159 / 0.689} & \textit{0.198 / 0.735} & \textit{0.213 / 0.764} & \textit{0.239 / \textbf{0.810}} & \textit{0.257 / 0.805} & \textit{0.242 / 0.805} \\
TTC$<$0.5$|$PET$<$1.0 (17.7) & 0.409 / 0.732 & 0.519 / \textbf{0.812} & 0.515 / 0.812 & 0.456 / 0.760 & 0.411 / 0.727 & 0.417 / 0.734 & 0.435 / 0.760 & 0.453 / 0.775 \\
TTC$<$1.0$|$PET$<$1.5 (37.1) & 0.620 / 0.754 & 0.739 / \textbf{0.864} & 0.731 / 0.855 & 0.673 / 0.809 & 0.639 / 0.772 & 0.646 / 0.776 & 0.664 / 0.797 & 0.702 / 0.840 \\
TTC$<$1.5$|$PET$<$2.0 (57.5) & 0.796 / 0.818 & 0.860 / \textbf{0.912} & 0.852 / 0.904 & 0.826 / 0.871 & 0.797 / 0.822 & 0.797 / 0.827 & 0.803 / 0.839 & 0.837 / 0.884 \\
\midrule
Average             & 0.472 / 0.765 & 0.537 / 0.826 & 0.525 / 0.815 & 0.504 / 0.801 & 0.477 / 0.778 & 0.493 / 0.801 & 0.509 / 0.810 & 0.525 / \textbf{0.835} \\
\midrule
\multicolumn{9}{c}{\textbf{US-101 Dataset}} \\
TTC$<$0.5 (5.6)   & 0.232 / 0.809 & 0.472 / 0.928 & 0.547 / \textbf{0.952} & 0.428 / 0.914 & 0.288 / 0.814 & 0.281 / 0.822 & 0.373 / 0.874 & 0.248 / 0.774 \\
TTC$<$1.0 (13.7)  & 0.428 / 0.793 & 0.690 / 0.944 & 0.713 / \textbf{0.949} & 0.650 / 0.933 & 0.469 / 0.834 & 0.472 / 0.846 & 0.548 / 0.884 & 0.566 / 0.902 \\
TTC$<$1.5 (27.4)  & 0.608 / 0.813 & 0.807 / \textbf{0.947} & 0.793 / 0.940 & 0.746 / 0.923 & 0.637 / 0.840 & 0.655 / 0.863 & 0.650 / 0.861 & 0.706 / 0.899 \\
\rowcolor{gray!15}
\textit{TTC Avg}    & \textit{0.423 / 0.805} & \textit{0.656 / \textbf{0.940}} & \textit{0.684 / 0.947} & \textit{0.608 / 0.923} & \textit{0.465 / 0.829} & \textit{0.469 / 0.844} & \textit{0.524 / 0.873} & \textit{0.507 / 0.858} \\
PET$<$1.0 (3.5)   & 0.166 / 0.757 & 0.151 / 0.728 & 0.131 / 0.687 & 0.154 / 0.758 & 0.168 / 0.777 & 0.235 / 0.859 & 0.225 / 0.834 & 0.256 / \textbf{0.875} \\
PET$<$1.5 (4.2)   & 0.194 / 0.768 & 0.144 / 0.735 & 0.149 / 0.718 & 0.190 / 0.768 & 0.161 / 0.774 & 0.254 / \textbf{0.880} & 0.228 / 0.850 & 0.276 / 0.870 \\
PET$<$2.0 (4.4)   & 0.212 / 0.790 & 0.197 / 0.773 & 0.177 / 0.753 & 0.221 / 0.779 & 0.202 / 0.769 & 0.267 / 0.855 & 0.295 / \textbf{0.874} & 0.285 / 0.873 \\
\rowcolor{gray!15}
\textit{PET Avg}    & \textit{0.191 / 0.772} & \textit{0.164 / 0.745} & \textit{0.152 / 0.719} & \textit{0.188 / 0.768} & \textit{0.177 / 0.773} & \textit{0.252 / 0.865} & \textit{0.249 / 0.853} & \textit{0.272 / \textbf{0.873}} \\
TTC$<$0.5$|$PET$<$1.0 (8.9) & 0.296 / 0.771 & 0.457 / \textbf{0.849} & 0.423 / 0.831 & 0.352 / 0.798 & 0.275 / 0.735 & 0.372 / 0.841 & 0.347 / 0.805 & 0.385 / 0.840 \\
TTC$<$1.0$|$PET$<$1.5 (17.3) & 0.493 / 0.798 & 0.666 / \textbf{0.900} & 0.658 / 0.897 & 0.576 / 0.870 & 0.509 / 0.812 & 0.526 / 0.837 & 0.558 / 0.863 & 0.609 / 0.888 \\
TTC$<$1.5$|$PET$<$2.0 (30.3) & 0.614 / 0.800 & 0.787 / \textbf{0.918} & 0.762 / 0.910 & 0.724 / 0.891 & 0.635 / 0.811 & 0.652 / 0.830 & 0.676 / 0.859 & 0.710 / 0.881 \\
\midrule
Average             & 0.360 / 0.789 & 0.486 / 0.858 & 0.484 / 0.849 & 0.449 / 0.848 & 0.372 / 0.796 & 0.413 / 0.848 & 0.433 / 0.856 & 0.449 / \textbf{0.867} \\
\bottomrule
\end{tabular}
\end{table*}


\subsection{Evaluation Metrics}
We evaluate all models using two complementary metrics. The primary metric is the Area Under the Receiver Operating Characteristic Curve (AUC), which measures the model's ability to rank frames by risk severity independently of any classification threshold. AUC is particularly suited to traffic conflict prediction because the operational goal is risk prioritization — identifying which frames warrant attention — rather than binary classification at a fixed threshold. Moreover, AUC is robust to class imbalance, which varies substantially across our nine configurations (from 3.5\% to 57.5\% risk prevalence).
As a secondary metric, we report the threshold-optimized F1 score, where the classification threshold is selected to maximize F1 on the validation set. F1 provides a measure of the best achievable classification performance but is sensitive to the probability distribution shape and class balance, making it less stable for cross-configuration comparison.

\subsection{Overall Comparison}
Table~\ref{tab:performance} present the full comparison of all eight methods across the nine threshold configurations on the I-80 and US-101 datasets, respectively. Two clear patterns emerge from the aggregate results. First, the proposed HIA-GAT achieves the highest average AUC on both datasets (0.835 on I-80, 0.867 on US-101), outperforming all non-graph and graph baselines. Second, Random Forest achieves the highest average F1 (0.537 on I-80, 0.486 on US-101). This divergence between AUC and F1 rankings reflects a fundamental distinction: AUC evaluates the full ranking of risk scores across all operating points, whereas F1 evaluates classification accuracy at a single best threshold. HIA-GAT produces well-calibrated probability scores that capture nuanced risk gradations, while tree-based ensembles produce sharper probability distributions that yield cleaner decision boundaries at optimal thresholds despite slightly inferior overall ranking.
Among the GNN baselines, a consistent architectural progression is observed. GCN, which operates on graph topology alone without edge features, achieves the lowest GNN performance (average AUC of 0.778 on I-80, 0.796 on US-101). GraphSAGE improves upon GCN through its neighborhood sampling mechanism but similarly lacks edge feature integration. HomoGAT, which incorporates 3-dimensional physics-informed edge features into the attention computation, yields a substantial improvement over GCN (+0.032 AUC on I-80, +0.060 on US-101), demonstrating that edge features encoding pairwise vehicle dynamics are critical for risk prediction. The proposed HIA-GAT extends this further with its dual-stream architecture and event-level gate supervision, achieving +0.025 AUC over HomoGAT on I-80 and +0.011 on US-101, and winning 7 of 9 configurations on both datasets by AUC.
\subsection{Conflict-Type Stratified Analysis}
The nine threshold configurations can be partitioned into three categories — TTC-only (rear-end conflicts), PET-only (lane-change conflicts), and combined — revealing distinct patterns in model suitability.
On TTC-only configurations, non-graph models perform strongly due to the nature of rear-end conflict detection. TTC is fundamentally determined by the closing rate and gap distance between a following vehicle and its leader — quantities that are well captured by aggregate frame-level statistics such as the mean and maximum closing rate across all vehicle pairs. Random Forest achieves average TTC AUC of 0.901 on I-80 and 0.940 on US-101. GNN models remain competitive (HIA-GAT: 0.868 and 0.858) but the additional representational capacity of graph-level reasoning provides diminishing returns when the risk signal is already well-encoded in simple summary statistics.
On PET-only configurations, the pattern reverses decisively. Non-graph methods degrade substantially, with RF and XGBoost achieving average PET AUC of only 0.715 and 0.689 on I-80, respectively (Table \ref{tab:performance}). In contrast, all GNN models outperform all non-graph baselines, with HIA-GAT and GraphSAGE achieving 0.805 and 0.810 on I-80, and 0.873 and 0.865 on US-101. This gap — approximately 0.10 AUC separating graph from non-graph methods — provides strong evidence that graph structure is essential for lane-change conflict detection. PET conflicts are inherently relational events: they arise from the spatial interaction between a lane-changing vehicle and an adjacent-lane occupant sharing longitudinal road space. Aggregate statistics collapse this pairwise structure, losing the specific vehicle pair information that makes PET detection possible.
On combined configurations (TTC|PET), the results reflect an intermediate regime. RF leads overall due to its strong TTC performance, but HIA-GAT achieves the largest improvement over HomoGAT in this category (+0.034 AUC on I-80, +0.028 on US-101), suggesting that the dual-stream gating mechanism provides its greatest value when both conflict types coexist within the same frame and must be differentiated. The exception is $\mathrm{TTC}<0.5$ on US-101 (5.6\% prevalence), where HIA-GAT underperforms on both F1 and AUC due to training instability on this extremely sparse configuration --- the dual-stream architecture splits an already limited set of positive examples across two streams, reducing the effective supervision signal per stream below the threshold needed for stable attention learning.

\subsection{Gate Supervision Analysis}
A central design feature of HIA-GAT is the conflict-type aware gating mechanism. Table~\ref{tab:gate} summarizes the gate behavior across all nine configurations on both datasets. We evaluate whether the gate successfully learns to differentiate between longitudinal and lateral conflicts through its supervision signal. On single-type configurations, the gate achieves near-perfect conflict type routing. For TTC-only configurations, the average gate value across all nodes converges above 0.5 (0.69-0.77 on I-80, 0.58--0.67 on US-101), indicating that the majority of the representation is routed through the longitudinal stream. For PET-only configurations, the gate converges below 0.5 (0.43--0.46 on I-80, 0.40--0.43 on US-101), routing the representation predominantly through the lateral stream. The gate direction accuracy -- defined as the fraction of supervised nodes where the gate value correctly exceeds 0.5 for TTC-involved vehicles or falls below 0.5 for PET-involved vehicles --- exceeds 99\% on all single-type configurations across both datasets.
\begin{figure}
    \centering
    \includegraphics[width=1\linewidth]{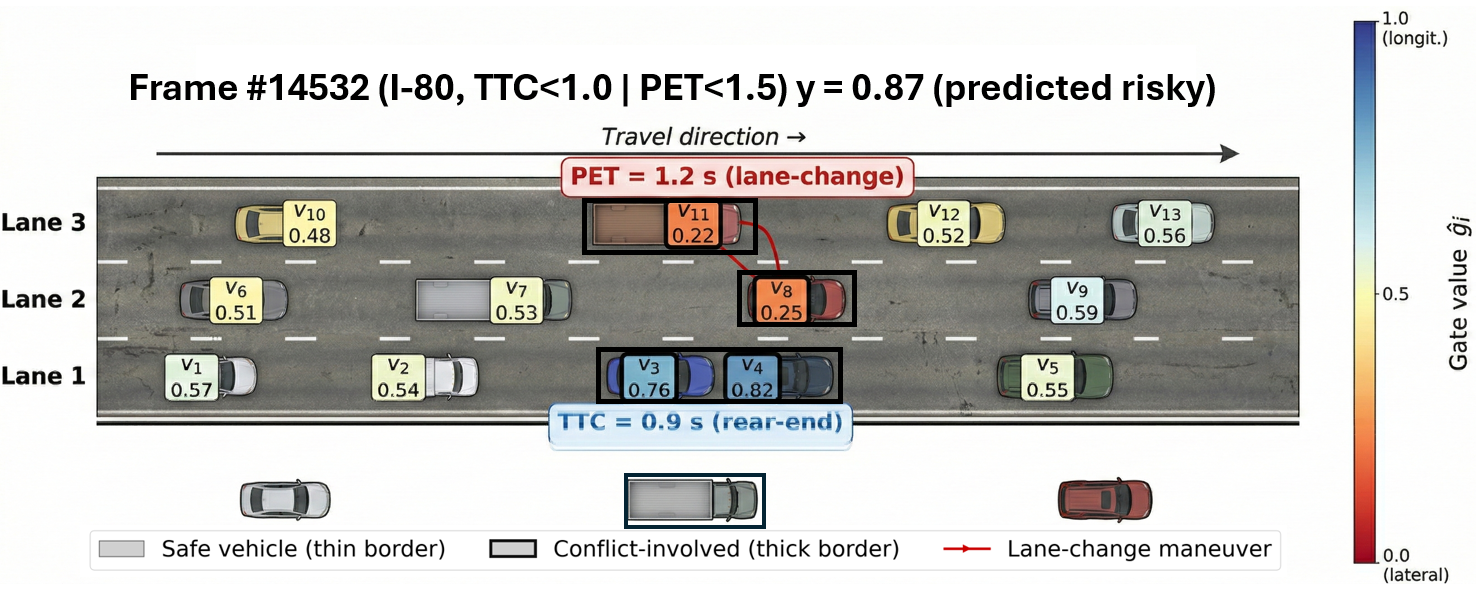}
    \caption{Per-vehicle gate values on a representative I-80 test frame (TTC$<$1.0\,|\,PET$<$1.5). Blue vehicles ($\bar{g}_i \to 1.0$) are routed through the longitudinal stream; red vehicles ($\bar{g}_i \to 0.0$) through the lateral stream. The gate correctly identifies the rear-end conflict pair ($v_3$/$v_4$) and the lane-change conflict pair ($v_8$/$v_{11}$), while safe vehicles remain near 0.5.}
    \label{fig:gate}
\end{figure}
On combined configurations, the gate exhibits substantial per-node variation (standard deviation 0.19--0.29), indicating that different vehicles within the same frame are routed through different streams depending on their conflict involvement. The aggregate gate mean shifts toward the TTC-dominant direction (0.54--0.70) because TTC events are approximately 6--11 times more numerous than PET events, which reduces PET gate direction accuracy on the most imbalanced combined configurations. Notably, only 0.1--1.5\% of nodes in each frame receive gate supervision, yet the gate generalizes to all nodes through the shared projection layer --- demonstrating that the model learns a mapping from node representations to conflict-type likelihood rather than memorizing supervised examples.

\begin{table}[t]
\centering
\footnotesize
\setlength{\tabcolsep}{2.5pt}
\caption{Gate supervision analysis for HIA-GAT. $\bar{g}$: mean gate value across all nodes (1.0 = fully longitudinal, 0.0 = fully lateral). $\sigma_g$: standard deviation. Dir.\ Acc.: fraction of supervised nodes with correct gate direction ($>$0.5 for TTC, $<$0.5 for PET).}
\label{tab:gate}
\begin{tabular}{l cc c cc c}
\toprule
 & \multicolumn{3}{c}{\textbf{I-80}} & \multicolumn{3}{c}{\textbf{US-101}} \\
\cmidrule(lr){2-4} \cmidrule(lr){5-7}
Configuration & $\bar{g}$ & $\sigma_g$ & Dir.\ Acc. & $\bar{g}$ & $\sigma_g$ & Dir.\ Acc. \\
\midrule
\multicolumn{7}{l}{\textit{TTC-only (expected: $\bar{g} > 0.5$)}} \\
TTC$<$0.5   & 0.71 & 0.08 & 100\%  & 0.67 & 0.23 & 99.7\% \\
TTC$<$1.0   & 0.69 & 0.09 & 100\%  & 0.58 & 0.24 & 99.9\% \\
TTC$<$1.5   & 0.77 & 0.07 & 100\%  & 0.60 & 0.25 & 100\%  \\
\midrule
\multicolumn{7}{l}{\textit{PET-only (expected: $\bar{g} < 0.5$)}} \\
PET$<$1.0   & 0.46 & 0.12 & 99.9\% & 0.40 & 0.12 & 99.1\% \\
PET$<$1.5   & 0.46 & 0.13 & 99.9\% & 0.41 & 0.15 & 100\%  \\
PET$<$2.0   & 0.43 & 0.14 & 100\%  & 0.43 & 0.15 & 100\%  \\
\midrule
\multicolumn{7}{l}{\textit{Combined (expected: per-node variation)}} \\
TTC$<$0.5$|$PET$<$1.0 & 0.60 & 0.23 & 89--91\% & 0.54 & 0.25 & 69--77\% \\
TTC$<$1.0$|$PET$<$1.5 & 0.64 & 0.25 & 88--93\% & 0.55 & 0.27 & 77--99\% \\
TTC$<$1.5$|$PET$<$2.0 & 0.70 & 0.19 & 80--95\% & 0.56 & 0.29 & 18--99\% \\
\bottomrule
\end{tabular}
\end{table}

\subsection{Discussion}
The most consequential finding of this study is the stark performance gap between graph-based and non-graph methods on PET-only configurations. While non-graph baselines achieve competitive or superior performance on TTC detection, they consistently fail on PET detection, with AUC values in the 0.69–0.77 range compared to 0.80–0.87 for GNN models. This finding has a clear physical interpretation. Rear-end conflicts are characterized by aggregate kinematic signatures — high mean closing rates, small minimum headways — that are preserved by statistical summaries of frame-level features. Lane-change conflicts, by contrast, are inherently pairwise events defined by the spatial and temporal relationship between two specific vehicles: one changing lanes and one occupying the target space. When these pairwise interactions are collapsed into frame-level means and maxima, the critical relational information is lost.
This result suggests that graph-based representations are not merely a modeling convenience for traffic risk prediction, but a necessary structural choice for comprehensive conflict detection that spans both longitudinal and lateral interaction types.

The comparison between GCN/GraphSAGE and HomoGAT/HIA-GAT isolates the value of physics-informed edge features. On I-80 TTC configurations, HomoGAT improves over GCN by $+0.030$ average AUC by explicitly encoding closing rate, gap distance, and acceleration differential—the key variables underlying TTC. HIA-GAT further extends this idea through conflict-specific streams: the longitudinal stream uses closing rate and acceleration differential, while the lateral stream uses lateral velocity, lane-change flag, and longitudinal overlap. This prevents one attention function from learning two distinct conflict mechanisms simultaneously. The gate also improves interpretability by indicating the dominant conflict type for each risky vehicle: values near 1.0 correspond to TTC-related rear-end interactions, values near 0.0 to PET-related lane-change conflicts, and safe vehicles remain near 0.5, as shown in Figure~\ref{fig:gate}. This attribution is obtained from SSM-derived labels with negligible added annotation or computational cost.

\section{VI Conclusion}

This study proposed HIA-GAT, a heterogeneous interaction-aware graph attention network for freeway traffic conflict risk prediction that processes longitudinal and lateral vehicle interactions through dedicated dual-stream pathways with physics-informed edge features and a conflict-type-aware gating mechanism. Evaluated against seven baselines across nine threshold configurations on the NGSIM I-80 and US-101 datasets, HIA-GAT achieved the highest average AUC on both datasets (0.835 and 0.867), with event-level gate supervision attaining over 99\% conflict-type routing accuracy on single-type configurations. The comprehensive comparison revealed that graph structure is essential for lane-change conflict detection, where GNN models outperformed non-graph baselines by approximately 0.10 AUC, while physics-informed edge features proved critical for rear-end conflict prediction. Beyond predictive performance, the gate mechanism provides per-vehicle interpretability — identifying which vehicles are involved in which conflict type — offering actionable information for traffic safety monitoring systems. 

\section{VII Acknowledgement}

This work was supported in part by the Transportation Network Growth Opportunity (TNGO) initiative funded by the Tennessee Department of Economic and Community Development, in collaboration with the University of Tennessee at Chattanooga and industry partners.

\bibliographystyle{unsrt}  


\begin{thebibliography}{1}

\bibitem{nhtsa_fars}
National Highway Traffic Safety Administration.
\newblock Fatality Analysis Reporting System (FARS).
\newblock \url{https://www.nhtsa.gov/research-data/fatality-analysis-reporting-system-fars}, n.d., Accessed: 2025-07-07.

\bibitem{fhwa_intersection_safety}
Federal Highway Administration.
\newblock About Intersection Safety.
\newblock \url{https://highways.dot.gov/safety/intersection-safety/about}, n.d., Accessed: 2025-07-07.

\bibitem{tay2007factors}
Richard Tay and Shakil Mohammad Rifaat.
\newblock Factors contributing to the severity of intersection crashes.
\newblock {\em Journal of Advanced Transportation}, 41(3):245--265, 2007.

\bibitem{wang2006temporal}
Xuesong Wang and Mohamed Abdel-Aty.
\newblock Temporal and spatial analyses of rear-end crashes at signalized intersections.
\newblock {\em Accident Analysis \& Prevention}, 38(6):1137--1150, 2006.

\bibitem{wang2007right}
Xuesong Wang and Mohamed Abdel-Aty.
\newblock Right-angle crash occurrence at signalized intersections.
\newblock {\em Transportation Research Record}, 2019(1):156--168, 2007.

\bibitem{northmore2019intersection}
Andrew Northmore and Eric Hildebrand.
\newblock Intersection characteristics that influence collision severity and cost.
\newblock {\em Journal of safety research}, 70:49--57, 2019.

\bibitem{turner1998intersection}
Shane Turner and Alan Nicholson.
\newblock Intersection accident estimation: the role of intersection location and non-collision flows.
\newblock {\em Accident Analysis \& Prevention}, 30(4):505--517, 1998.

\bibitem{salim2007collision}
Flora Dilys Salim, Seng Wai Loke, Andry Rakotonirainy, Bala Srinivasan, and Shonali Krishnaswamy.
\newblock Collision pattern modeling and real-time collision detection at road intersections.
\newblock In {\em 2007 IEEE Intelligent transportation systems conference}, pages 161--166. IEEE, 2007.

\bibitem{sarkar2021role}
Abhijit Sarkar, Hananeh Alambeigi, Anthony McDonald, Gustav Markkula, and Jeff Hickman.
\newblock Role of peripheral vision in brake reaction time during safety critical events.
\newblock In {\em Proceedings of the Human Factors and Ergonomics Society Annual Meeting}, volume 65, number 1, pages 695--699. SAGE Publications Sage CA: Los Angeles, CA, 2021.

\bibitem{liu2007association}
Bor-Shong Liu.
\newblock Association of intersection approach speed with driver characteristics, vehicle type and traffic conditions comparing urban and suburban areas.
\newblock {\em Accident Analysis \& Prevention}, 39(2):216--223, 2007.

\bibitem{larsen2002multidisciplinary}
Lotte Larsen and Peter Kines.
\newblock Multidisciplinary in-depth investigations of head-on and left-turn road collisions.
\newblock {\em Accident Analysis \& Prevention}, 34(3):367--380, 2002.

\bibitem{chitraranjan2025vision}
Charith Chitraranjan, Vipooshan Vipulananthan, and Thuvarakan Sritharan.
\newblock Vision-Based Collision Warning Systems with Deep Learning: A Systematic Review.
\newblock {\em Journal of Imaging}, 11(2):64, 2025.

\bibitem{razi2023deep}
Abolfazl Razi, Xiwen Chen, Huayu Li, Hao Wang, Brendan Russo, Yan Chen, and Hongbin Yu.
\newblock Deep learning serves traffic safety analysis: A forward-looking review.
\newblock {\em IET Intelligent Transport Systems}, 17(1):22--71, 2023.

\bibitem{mahmud2019micro}
SM Sohel Mahmud, Luis Ferreira, Md Shamsul Hoque, and Ahmad Tavassoli.
\newblock Micro-simulation modelling for traffic safety: A review and potential application to heterogeneous traffic environment.
\newblock {\em IATSS research}, 43(1):27--36, 2019.

\bibitem{lee2014development}
Jaeyoung Lee, Mohamed Abdel-Aty, and Ximiao Jiang.
\newblock Development of zone system for macro-level traffic safety analysis.
\newblock {\em Journal of transport geography}, 38:13--21, 2014.

\bibitem{abdelwahab2002artificial}
Hassan T Abdelwahab and Mohamed A Abdel-Aty.
\newblock Artificial neural networks and logit models for traffic safety analysis of toll plazas.
\newblock {\em Transportation Research Record}, 1784(1):115--125, 2002.

\bibitem{huang2010multilevel}
Helai Huang and Mohamed Abdel-Aty.
\newblock Multilevel data and Bayesian analysis in traffic safety.
\newblock {\em Accident Analysis \& Prevention}, 42(6):1556--1565, 2010.

\bibitem{murat2017integration}
Yetis Sazi Murat and Ziya Cakici.
\newblock An integration of different computing approaches in traffic safety analysis.
\newblock {\em Transportation research procedia}, 22:265--274, 2017.

\bibitem{nguyen2022assessment}
Tuan Thanh Nguyen and Phuong Thao Cao.
\newblock Assessment of Traffic Safety Between Pedestrians and Vehicles Using Traffic Conflict Technique.
\newblock In {\em International Conference on Sustainability in Civil Engineering}, pages 763--769. Springer, 2022.

\bibitem{paul2020post}
Madhumita Paul and Indrajit Ghosh.
\newblock Post encroachment time threshold identification for right-turn related crashes at unsignalized intersections on intercity highways under mixed traffic.
\newblock {\em International journal of injury control and safety promotion}, 27(2):121--135, 2020.

\bibitem{qi2020modified}
Weiwei Qi, Wei Wang, Bin Shen, and Jiabin Wu.
\newblock A modified post encroachment time model of urban road merging area based on lane-change characteristics.
\newblock {\em IEEE Access}, 8:72835--72846, 2020.

\bibitem{sonth2023real}
Akash Sonth, Abhijit Sarkar, Sparsh Jain, Hirva Bhagat, and Zachary R Doerzaph.
\newblock Real-time risk prediction at signalized intersections using a graph neural network.
\newblock Safe-D University Transportation Center, 2023.

\bibitem{traffic2009manual}
C Traffic.
\newblock Manual on uniform traffic control devices.
\newblock {\em US Department of Transportation, Federal Highway Administation}, 2009.

\bibitem{gershon2019distracted}
Pnina Gershon, Kellienne R Sita, Chunming Zhu, Johnathon P Ehsani, Sheila G Klauer, Tom A Dingus, and Bruce G Simons-Morton.
\newblock Distracted driving, visual inattention, and crash risk among teenage drivers.
\newblock {\em American journal of preventive medicine}, 56(4):494--500, 2019.

\bibitem{lord2010statistical}
Dominique Lord and Fred Mannering.
\newblock The statistical analysis of crash-frequency data: A review and assessment of methodological alternatives.
\newblock {\em Transportation research part A: policy and practice}, 44(5):291--305, 2010.

\bibitem{savolainen2011statistical}
Peter T Savolainen, Fred L Mannering, Dominique Lord, and Mohammed A Quddus.
\newblock The statistical analysis of highway crash-injury severities: A review and assessment of methodological alternatives.
\newblock {\em Accident Analysis \& Prevention}, 43(5):1666--1676, 2011.

\bibitem{huang2008severity}
Helai Huang, Hoong Chor Chin, and Md Mazharul Haque.
\newblock Severity of driver injury and vehicle damage in traffic crashes at intersections: a Bayesian hierarchical analysis.
\newblock {\em Accident Analysis \& Prevention}, 40(1):45--54, 2008.

\bibitem{billah2021analysis}
Khondoker Billah, Qasim Adegbite, Hatim O Sharif, Samer Dessouky, and Lauren Simcic.
\newblock Analysis of intersection traffic safety in the city of San Antonio, 2013--2017.
\newblock {\em Sustainability}, 13(9):5296, 2021.

\bibitem{hasain2022safety}
N Mohamed Hasain and Mokaddes Ali Ahmed.
\newblock Safety evaluation of unsignalized intersection with heterogeneous traffic using Post Encroachment Time and conflicting vehicle speed.
\newblock {\em European Transport/Trasporti Europei}, 88(88):1--14, 2022.

\bibitem{johnsson2021validation}
Carl Johnsson, Aliaksei Laureshyn, and Carmelo D{\'a}gostino.
\newblock Validation of surrogate measures of safety with a focus on bicyclist--motor vehicle interactions.
\newblock {\em Accident Analysis \& Prevention}, 153:106037, 2021.

\bibitem{vogel2003comparison}
Katja Vogel.
\newblock A comparison of headway and time to collision as safety indicators.
\newblock {\em Accident analysis \& prevention}, 35(3):427--433, 2003.

\bibitem{kiefer2006time}
Raymond J Kiefer, Carol A Flannagan, and Christian J Jerome.
\newblock Time-to-collision judgments under realistic driving conditions.
\newblock {\em Human factors}, 48(2):334--345, 2006.

\bibitem{brown2005adjusted}
Timothy L Brown.
\newblock Adjusted minimum time-to-collision (TTC): A robust approach to evaluating crash scenarios.
\newblock In {\em Proceedings of the Driving Simulation Conference North America}, volume 40, pages 40--48, 2005.

\bibitem{peesapati2018can}
Lakshmi N Peesapati, Michael P Hunter, and Michael O Rodgers.
\newblock Can post encroachment time substitute intersection characteristics in crash prediction models?
\newblock {\em Journal of safety research}, 66:205--211, 2018.

\bibitem{fu2021comparison}
Chuanyun Fu and Tarek Sayed.
\newblock Comparison of threshold determination methods for the deceleration rate to avoid a crash (DRAC)-based crash estimation.
\newblock {\em Accident Analysis \& Prevention}, 153:106051, 2021.

\bibitem{fadilah2014time}
Suzi Iryanti Fadilah and Azizul Rahman Mohd Shariff.
\newblock A time gap interval for safe following distance (TGFD) in avoiding car collision in wireless vehicular networks (VANET) environment.
\newblock In {\em 2014 5th International Conference on Intelligent Systems, Modelling and Simulation}, pages 683--689. IEEE, 2014.

\bibitem{minderhoud2001extended}
Michiel M Minderhoud and Piet HL Bovy.
\newblock Extended time-to-collision measures for road traffic safety assessment.
\newblock {\em Accident Analysis \& Prevention}, 33(1):89--97, 2001.

\bibitem{hayward1972near}
John C Hayward.
\newblock Near miss determination through use of a scale of danger.
\newblock Pennsylvania State University University Park, 1972.

\bibitem{bisulco2025ev}
Anthony Bisulco, Vijay Kumar, and Kostas Daniilidis.
\newblock EV-TTC: Event-Based Time to Collision under Low Light Conditions.
\newblock {\em IEEE Robotics and Automation Letters}, 2025.

\bibitem{therattil2025safety}
Jino Thomas Therattil, More Prathamesh Avinash, and Nipjyoti Bharadwaj.
\newblock Safety Analysis at Unsignalized T Intersection Using PET and Extreme Value Theorem.
\newblock {\em Transportation in Developing Economies}, 11(1):1--15, 2025.

\bibitem{memory1997sepp}
Long Short-Term Memory.
\newblock Sepp hochreiter and j{\"u}rgen schmidhuber.
\newblock {\em Neural Computation}, 9(8):1735, 1997.

\bibitem{chen2017surrogate}
Peng Chen, Weiliang Zeng, Guizhen Yu, and Yunpeng Wang.
\newblock Surrogate safety analysis of pedestrian-vehicle conflict at intersections using unmanned aerial vehicle videos.
\newblock {\em Journal of advanced transportation}, 2017(1):5202150, 2017.

\bibitem{yu2019traffic}
Quan Yu and Yuting Zhou.
\newblock Traffic safety analysis on mixed traffic flows at signalized intersection based on Haar-Adaboost algorithm and machine learning.
\newblock {\em Safety Science}, 120:248--253, 2019.

\bibitem{kucskapan2022pedestrian}
Emre Ku{\c{s}}kapan, Mohammad Ali Sahraei, Merve Kayaci {\c{C}}odur, and Muhammed Yasin {\c{C}}odur.
\newblock Pedestrian safety at signalized intersections: Spatial and machine learning approaches.
\newblock {\em Journal of Transport \& Health}, 24:101322, 2022.

\bibitem{silva2020machine}
Philippe Barbosa Silva, Michelle Andrade, and Sara Ferreira.
\newblock Machine learning applied to road safety modeling: A systematic literature review.
\newblock {\em Journal of traffic and transportation engineering (English edition)}, 7(6):775--790, 2020.

\bibitem{zyner2018recurrent}
Alex Zyner, Stewart Worrall, and Eduardo Nebot.
\newblock A recurrent neural network solution for predicting driver intention at unsignalized intersections.
\newblock {\em IEEE Robotics and Automation Letters}, 3(3):1759--1764, 2018.

\bibitem{lu2024detection}
Lingxin Lu, Hui Wang, Yan Wan, and Feifei Xu.
\newblock A detection transformer-based intelligent identification method for multiple types of road traffic safety facilities.
\newblock {\em Sensors}, 24(10):3252, 2024.

\bibitem{gomes2017embedded}
Samuel L Gomes, Eliz{\^a}ngela de S Rebou{\c{c}}as, Edson Cavalcanti Neto, Jo{\~a}o P Papa, Victor HC de Albuquerque, Pedro P Rebou{\c{c}}as Filho, and Joao Manuel RS Tavares.
\newblock Embedded real-time speed limit sign recognition using image processing and machine learning techniques.
\newblock {\em Neural Computing and Applications}, 28(Suppl 1):573--584, 2017.

\bibitem{neto2015brazilian}
Edson Cavalcanti Neto, Samuel Luz Gomes, Pedro Pedrosa Rebou{\c{c}}as Filho, and Victor Hugo C de Albuquerque.
\newblock Brazilian vehicle identification using a new embedded plate recognition system.
\newblock {\em Measurement}, 70:36--46, 2015.

\bibitem{scholl2019surrogate}
Lynn Scholl, Mohamed Elagaty, Bismarck Ledezma-Navarro, Edgar Zamora, and Luis Miranda-Moreno.
\newblock A surrogate video-based safety methodology for diagnosis and evaluation of low-cost pedestrian-safety countermeasures: The case of Cochabamba, Bolivia.
\newblock {\em Sustainability}, 11(17):4737, 2019.

\bibitem{nippani2023graph}
Abhinav Nippani, Dongyue Li, Haotian Ju, Haris Koutsopoulos, and Hongyang Zhang.
\newblock Graph neural networks for road safety modeling: Datasets and evaluations for accident analysis.
\newblock {\em Advances in neural information processing systems}, 36:52009--52032, 2023.

\bibitem{kipf2016semi}
Thomas N Kipf and Max Welling.
\newblock Semi-supervised classification with graph convolutional networks.
\newblock {\em arXiv preprint arXiv:1609.02907}, 2016.

\bibitem{wu2018moleculenet}
Zhenqin Wu, Bharath Ramsundar, Evan N Feinberg, Joseph Gomes, Caleb Geniesse, Aneesh S Pappu, Karl Leswing, and Vijay Pande.
\newblock MoleculeNet: a benchmark for molecular machine learning.
\newblock {\em Chemical science}, 9(2):513--530, 2018.

\bibitem{scarselli2008graph}
Franco Scarselli, Marco Gori, Ah Chung Tsoi, Markus Hagenbuchner, and Gabriele Monfardini.
\newblock The graph neural network model.
\newblock {\em IEEE transactions on neural networks}, 20(1):61--80, 2008.

\bibitem{mo2021heterogeneous}
Xiaoyu Mo, Yang Xing, and Chen Lv.
\newblock Heterogeneous edge-enhanced graph attention network for multi-agent trajectory prediction.
\newblock {\em arXiv preprint arXiv:2106.07161}, 2021.

\bibitem{malawade2022spatiotemporal}
Arnav Vaibhav Malawade, Shih-Yuan Yu, Brandon Hsu, Deepan Muthirayan, Pramod P Khargonekar, and Mohammad Abdullah Al Faruque.
\newblock Spatiotemporal scene-graph embedding for autonomous vehicle collision prediction.
\newblock {\em IEEE Internet of Things Journal}, 9(12):9379--9388, 2022.

\bibitem{diehl2019graph}
Frederik Diehl, Thomas Brunner, Michael Truong Le, and Alois Knoll.
\newblock Graph neural networks for modelling traffic participant interaction.
\newblock In {\em 2019 IEEE Intelligent Vehicles Symposium (IV)}, pages 695--701. IEEE, 2019.

\bibitem{oberoi2017spatial}
Kamaldeep Singh Oberoi, G{\'e}raldine Del Mondo, Yohan Dupuis, and Pascal Vasseur.
\newblock Spatial modeling of urban road traffic using graph theory.
\newblock In {\em Proceedings of Spatial Analysis and GEOmatics (SAGEO) 2017}, pages 264--277, 2017.

\bibitem{pan2025urban}
Yuyan Annie Pan et al.
\newblock Urban intersection traffic flow prediction: A physics-guided stepwise framework utilizing spatio-temporal graph neural network algorithms.
\newblock {\em Multimodal Transportation}, 4(2):100207, 2025.

\bibitem{fleck2019towards}
Tobias Fleck, Karam Daaboul, Michael Weber, Philip Sch{\"o}rner, Marek Wehmer, Jens Doll, Stefan Orf, Nico Su{\ss}mann, Christian Hubschneider, Marc Ren{\'e} Zofka et al.
\newblock Towards large scale urban traffic reference data: Smart infrastructure in the test area autonomous driving baden-w{\"u}rttemberg.
\newblock In {\em Intelligent Autonomous Systems 15: Proceedings of the 15th International Conference IAS-15}, pages 964--982. Springer, 2019.

\bibitem{zipfl2020traffic}
Maximilian Zipfl, Tobias Fleck, Marc Ren{\'e} Zofka, and J Marius Z{\"o}llner.
\newblock From traffic sensor data to semantic traffic descriptions: The test area autonomous driving baden-w{\"u}rttemberg dataset (taf-bw dataset).
\newblock In {\em 2020 IEEE 23rd International Conference on Intelligent Transportation Systems (ITSC)}, pages 1--7. IEEE, 2020.

\bibitem{bender2014lanelets}
Philipp Bender, Julius Ziegler, and Christoph Stiller.
\newblock Lanelets: Efficient map representation for autonomous driving.
\newblock In {\em 2014 IEEE Intelligent Vehicles Symposium Proceedings}, pages 420--425. IEEE, 2014.

\bibitem{poggenhans2018lanelet2}
Fabian Poggenhans, Jan-Hendrik Pauls, Johannes Janosovits, Stefan Orf, Maximilian Naumann, Florian Kuhnt, and Matthias Mayr.
\newblock Lanelet2: A High-Definition Map Framework for the Future of Automated Driving.
\newblock In {\em Proc.\ IEEE Intell.\ Trans.\ Syst.\ Conf.}, Hawaii, USA, November 2018. \url{http://www.mrt.kit.edu/z/publ/download/2018/Poggenhans2018Lanelet2.pdf}.

\bibitem{haklay2008openstreetmap}
Mordechai Haklay and Patrick Weber.
\newblock Openstreetmap: User-generated street maps.
\newblock {\em IEEE Pervasive computing}, 7(4):12--18, 2008.

\bibitem{shirazi2016looking}
Mohammad Shokrolah Shirazi and Brendan Tran Morris.
\newblock Looking at intersections: a survey of intersection monitoring, behavior and safety analysis of recent studies.
\newblock {\em IEEE Transactions on Intelligent Transportation Systems}, 18(1):4--24, 2016.

\bibitem{allen1978analysis}
Brian L Allen, B Tom Shin, and Peter J Cooper.
\newblock Analysis of traffic conflicts and collisions.
\newblock Technical report, 1978.

\bibitem{van2014traffic}
A Richard A Van der Horst, Maartje de Goede, Stefanie de Hair-Buijssen, and Rob Methorst.
\newblock Traffic conflicts on bicycle paths: A systematic observation of behaviour from video.
\newblock {\em Accident Analysis \& Prevention}, 62:358--368, 2014.

\bibitem{zhou2013development}
Huanyun Zhou and Fei Huang.
\newblock Development of traffic safety evaluation method based on simulated conflicts at signalized intersections.
\newblock {\em Procedia-Social and Behavioral Sciences}, 96:881--885, 2013.

\bibitem{zipfl2022towards}
Maximilian Zipfl and J Marius Z{\"o}llner.
\newblock Towards traffic scene description: The semantic scene graph.
\newblock In {\em 2022 IEEE 25th International Conference on Intelligent Transportation Systems (ITSC)}, pages 3748--3755. IEEE, 2022.

\bibitem{douglas2011weisfeiler}
Brendan L Douglas.
\newblock The weisfeiler-lehman method and graph isomorphism testing.
\newblock {\em arXiv preprint arXiv:1101.5211}, 2011.

\bibitem{xu2018powerful}
Keyulu Xu, Weihua Hu, Jure Leskovec, and Stefanie Jegelka.
\newblock How powerful are graph neural networks?
\newblock {\em arXiv preprint arXiv:1810.00826}, 2018.

\bibitem{velivckovic2017graph}
Petar Veli{\v{c}}kovi{\'c}, Guillem Cucurull, Arantxa Casanova, Adriana Romero, Pietro Lio, and Yoshua Bengio.
\newblock Graph attention networks.
\newblock {\em arXiv preprint arXiv:1710.10903}, 2017.

\bibitem{gilmer2020message}
Justin Gilmer, Samuel S Schoenholz, Patrick F Riley, Oriol Vinyals, and George E Dahl.
\newblock Message passing neural networks.
\newblock In {\em Machine learning meets quantum physics}, pages 199--214. Springer, 2020.

\bibitem{FHWA_NGSIM_I80_2016}
U.S. Department of Transportation Federal Highway Administration.
\newblock Next Generation Simulation (NGSIM) Program I-80 Videos.
\newblock \url{https://data.transportation.gov}, 2016, [Dataset]. Provided by ITS DataHub through Data.transportation.gov.

\bibitem{singh2024conflict}
Dungar Singh, Pritikana Das, and Indrajit Ghosh.
\newblock Conflict-Based safety evaluations at unsignalized intersections using surrogate safety measures.
\newblock {\em Heliyon}, 10(5), 2024.

\bibitem{bonela2022review}
Someswara Rao Bonela and B Raghuram Kadali.
\newblock Review of traffic safety evaluation at T-intersections using surrogate safety measures in developing countries context.
\newblock {\em IATSS research}, 46(3):307--321, 2022.

\bibitem{chaudhari2021exploring}
Avinash Chaudhari, Ninad Gore, Shriniwas Arkatkar, Gaurang Joshi, and Srinivas Pulugurtha.
\newblock Exploring pedestrian surrogate safety measures by road geometry at midblock crosswalks: A perspective under mixed traffic conditions.
\newblock {\em IATSS research}, 45(1):87--101, 2021.

\bibitem{bataineh2025evaluating}
Tamer Bataineh, Nischal Bhattarai, Keshav Jimee, Yibin Zhang, Martin Lucero, Junxuan Zhao, and Hongchao Liu.
\newblock Evaluating driving behavior and intersection safety using roadside LiDAR: A study of safety indicators across collision metrics.
\newblock {\em Journal of Transportation Safety \& Security}, 17(10):1105--1137, 2025.

\bibitem{hasain2024proposing}
N Mohamed Hasain and Mokaddes Ali Ahmed.
\newblock Proposing an effective approach for traffic safety assessment on heterogeneous traffic conditions using surrogate safety measures and speed of the involved vehicles.
\newblock {\em Traffic injury prevention}, 25(2):219--227, 2024.

\bibitem{dimitrijevic2022short}
Branislav Dimitrijevic, Sina Darban Khales, Roksana Asadi, and Joyoung Lee.
\newblock Short-term segment-level crash risk prediction using advanced data modeling with proactive and reactive crash data.
\newblock {\em Applied Sciences}, 12(2):856, 2022.

\bibitem{xiang2024research}
Wenjun Xiang, Chen Peng, Jian Feng, Tianyue Zhang, and Jinrui Zhu.
\newblock Research on the Lane-Changing Behavior of Drivers at Highway Interchanges.
\newblock In {\em CICTP 2024}, pages 2508--2517. 2024.

\bibitem{zhao2025traffic}
Yi Zhao, Chuwei Zhao, Ou Zheng, and Jianxiao Ma.
\newblock Traffic safety risk assessment and characterization of lane-changing behavior in urban expressway interchange weaving area.
\newblock {\em Travel Behaviour and Society}, 41:101106, 2025.

\bibitem{lin2025analyzing}
Yifeng Lin, Fengjuan Xu, and Huiying Wen.
\newblock Analyzing lane-changing characteristics when interacting with different current-lane preceding vehicles.
\newblock {\em Journal of Transportation Safety \& Security}, pages 1--36, 2025.

\bibitem{fhwa_using_safety_analyses_2025}
Federal Highway Administration (FHWA).
\newblock 5.0 Using Safety Analyses for Planning.
\newblock \url{https://highways.dot.gov/safety/zero-deaths/applying-safety-data-and-analysis-performance-based-transportation-planning/50}, 2025, U.S. Department of Transportation. Part of the guide 'Applying Safety Data and Analysis to Performance-Based Transportation Planning', Accessed: 2026-02-22.

\bibitem{ngsim_i80_2016}
U.S. Department of Transportation Federal Highway Administration.
\newblock Next Generation Simulation (NGSIM) Program I-80 Videos.
\newblock \url{https://doi.org/10.21949/1504477}, ITS DataHub through Data.transportation.gov, 2016.

\bibitem{ngsim_us101_2016}
U.S. Department of Transportation Federal Highway Administration.
\newblock Next Generation Simulation (NGSIM) Program US-101 Videos.
\newblock \url{https://doi.org/10.21949/1504477}, ITS DataHub through Data.transportation.gov, 2016.

\bibitem{peng2025pastgcn}
Fei Peng and HaiYang Xuan.
\newblock PASTGCN: A Real-Time Intersection Conflict Risk Prediction Model via Spatiotemporal Graph Convolutional Networks with Multimodal Sensor Fusion.
\newblock {\em Informatica}, 49(26), 2025.

\bibitem{muduli2026graph}
Kaliprasana Muduli, Indrajit Ghosh, and Satish V Ukkusuri.
\newblock A graph-based spatio-temporal framework for predicting safety-critical pedestrian--vehicle interactions at unsignalized crosswalks.
\newblock {\em Accident Analysis \& Prevention}, 228:108409, 2026.

\end{thebibliography}

\end{document}